%% file: main_arxiv.tex
\PassOptionsToPackage{table}{xcolor}
\documentclass{applemlr}

\input{apple_preamble}

\usepackage{float}

\newcommand{\method}{\textsc{DynaMiCS}\xspace}

\usepackage{xcolor}

\title{\textsc{DynaMiCS}: Fine-tuning LLMs with Performance Constraints using Dynamic Mixtures}

\author{Eleonora Gualdoni}
\author{Sonia Laguna}
\author{Louis B\'{e}thune}
\author{Joao Monteiro}
\author{Pierre Ablin}
\author{Marco Cuturi}

\affiliation{Apple}

\abstract{
Multi-domain fine-tuning of large language models requires improving performance on target domains while preserving previous capabilities, such as general knowledge, instruction following, or safety evaluations. Existing data mixing strategies rely on fixed heuristics or adaptive rules that cannot explicitly enforce preservation of such capabilities. We propose \method, a dynamic mixture optimizer that casts multi-domain fine-tuning as a constrained optimization problem. At each update, \method performs short domain-specific probing runs to estimate a slope matrix of local cross-domain effects, capturing how training on each fine-tuning dataset affects each evaluation domain. These estimates are then used to compute mixture weights through optimization over the probability simplex, with the objective of improving target-domain performance while keeping constrained-domain metrics within a specified tolerance of reference levels. Because these effects are measured by finite differences rather than gradients, targets and constraints need not be differentiable, or present in the fine-tuning data, and can be specified directly as benchmark accuracies. Across scenarios with varying numbers of target and constrained domains, and with loss- or accuracy-based objectives, \method achieves stronger target-domain improvements and higher constraint satisfaction than static, dynamic, similarity-based and probing-based alternatives, without a reference model, per-example scoring, or manually tuned weights.}

\metadata[Correspondence]{\sffamily Eleonora Gualdoni: \url{e_gualdoni@apple.com}, Sonia Laguna: \url{slaguna@ethz.ch}, Pierre Ablin: \url{p_ablin@apple.com}, Marco Cuturi: \url{m_cuturi@apple.com}}
\metadata[Note]{\sffamily SL contributed to this work while interning at Apple.}
\date{\sffamily\today}

\begin{document}

\maketitle

\input{sections/introduction}
\input{sections/related_work}

\input{sections/our_method}
\input{sections/experimental_setup}

\input{sections/results}

\input{sections/conclusion}

\bibliographystyle{plainnat}
\bibliography{main}

\newpage
\appendix

\input{sections/appendix}

\end{document}

%% file: apple_preamble.tex
\usepackage{amsmath}
\usepackage{enumerate}
\usepackage{algorithm}
\usepackage{algorithmic}
\usepackage{amsfonts}
\usepackage{amsthm}
\usepackage{amssymb}
\usepackage{xspace}
\usepackage{wrapfig}
\usepackage{adjustbox}
\usepackage{tabularx}
\usepackage{mathtools}
\usepackage{tikz}
\usepackage{enumitem}
\usepackage{silence}
\usepackage{dsfont}
\usepackage{multirow}
\usepackage{makecell}
\usepackage{xfakebold}
\usepackage{nicefrac}
\usepackage{pifont}
\input{math_commands}

\usepackage[table]{xcolor}
\usepackage{longtable}
\usepackage{booktabs}

\newcommand{\tinybm}{\ensuremath{\times}\text{tiny}}

\definecolor{grpA}{RGB}{237,243,252}  
\definecolor{grpB}{RGB}{245,245,245}  

\definecolor{textgray}{HTML}{6E6E73}
\usetikzlibrary{positioning, calc}
\usetikzlibrary{decorations.pathmorphing}

\makeatletter
\patchcmd{\wrong@fontshape}{\@gobbletwo}{}{}{}
\makeatother
\numberwithin{equation}{section}
\makeatletter
\AtBeginDocument{
  \urlstyle{sf}
  
}
\makeatother

\definecolor{light}{RGB}{125, 125, 125}
\crefname{tcb@cnt@pbox}{code}{code}
\Crefname{tcb@cnt@pbox}{Code}{Code}
\crefname{assumption}{assumption}{assumption}
\Crefname{assumption}{Assumption}{Assumptions}

\newtcolorbox[auto counter]{pbox}[2][]{
  colback=white,
  title=Code~\thetcbcounter: #2,
  #1,fonttitle=\sffamily,
  fontupper=\sffamily,
  arc=2pt,
  colframe=bgcolor,
  coltitle=fgcolor,
  colbacktitle=bgcolor,
  toptitle=0.25cm,
  bottomtitle=0.125cm
}

\makeatletter
\newcommand\applefootnote[1]{%
  \begingroup
  \renewcommand\thefootnote{}%
  \renewcommand\@makefntext[1]{\noindent##1}%
  \footnote{#1}%
  \addtocounter{footnote}{-1}%
  \endgroup
}
\makeatother

\definecolor{cverbbg}{gray}{0.90}

%% file: math_commands.tex
\usepackage{amsmath,amsfonts,bm}

\def\eqref#1{equation~\ref{#1}}

\def\1{\bm{1}}

\DeclareMathAlphabet{\mathsfit}{\encodingdefault}{\sfdefault}{m}{sl}
\SetMathAlphabet{\mathsfit}{bold}{\encodingdefault}{\sfdefault}{bx}{n}

%% file: sections/introduction.tex
\section{Introduction}

Fine-tuning large language models (LLMs) on new specialization domains presents a fundamental challenge: models must acquire new capabilities, such as code generation or mathematical reasoning, while retaining previously learned abilities, such as general knowledge, instruction following, or safety compliance.
This adaptation--retention tension is reflected in two related phenomena.
First, \textit{catastrophic forgetting}~\citep{mccloskey:catastrophic,Kirkpatrick_2017,shi2024continuallearninglargelanguage} degrades performance on domains previously mastered by the LLM.
Second, because supervised fine-tuning (SFT) datasets are typically small compared to pre-training corpora, large models tend to \textit{overfit}, adapting rapidly to the specialization domain while losing broader generalization abilities~\citep{shi2024instructiontuninglossinstructions, gupta2025selective, hao2025understanding}.
These risks motivate fine-tuning models with \emph{data mixtures}, where specialization data are interleaved with auxiliary or pre-training domains that can regularize adaptation and mitigate forgetting~\citep{he2021analyzingforgettingproblempretrainfinetuning, DBLP:journals/tmlr/IbrahimTGRABLR24,bethune2025pretraining_injection}.
\looseness-1

\textbf{What mixture for fine-tuning?} As models grow in size, post-training absorbs a larger share of total training computation, with SFT emerging as one of its key scaling dimensions~\citep{tie2025surveyposttraininglargelanguage, olmo2026olmo3}. 
In that context, the costs of picking suboptimal mixtures add up.
As a result, without a principled selection method, practitioners resort to trial-and-error attempts, where meaningfully covering the simplex grows exponentially as the number of fine-tuning datasets grows.
Simpler strategies such as uniform, dataset-size proportional, or temperature-based sampling are the most popular~\citep{wang2020balancingtrainingmultilingualneural, mueller-etal-2024-multi}.
These methods assume fixed inter-domain relationships throughout training and, crucially,
offer no way to guarantee that critical capabilities are retained.
More recently, a line of work updates mixtures during training, either from curriculum or loss-based signals~\citep{wu2024mixtureofskillslearningoptimizedata}, or by modeling cross-domain interactions and reweighting online~\citep{skillit, aioli}.
These methods adapt to evolving inter-domain relationships, but optimize unconstrained performance objectives: 
none lets a practitioner name a capability that must be preserved and state an explicit tolerance for its degradation.
Knowledge preservation is therefore treated only implicitly as part of the performance trade-off.
Finally, methods originally developed for pre-training rely on proxy models, %
making them costly to repurpose in the fine-tuning setting~\citep{doremi, doge, tandem}.

\begin{figure}[t] 
    \centering \includegraphics[width=\linewidth]{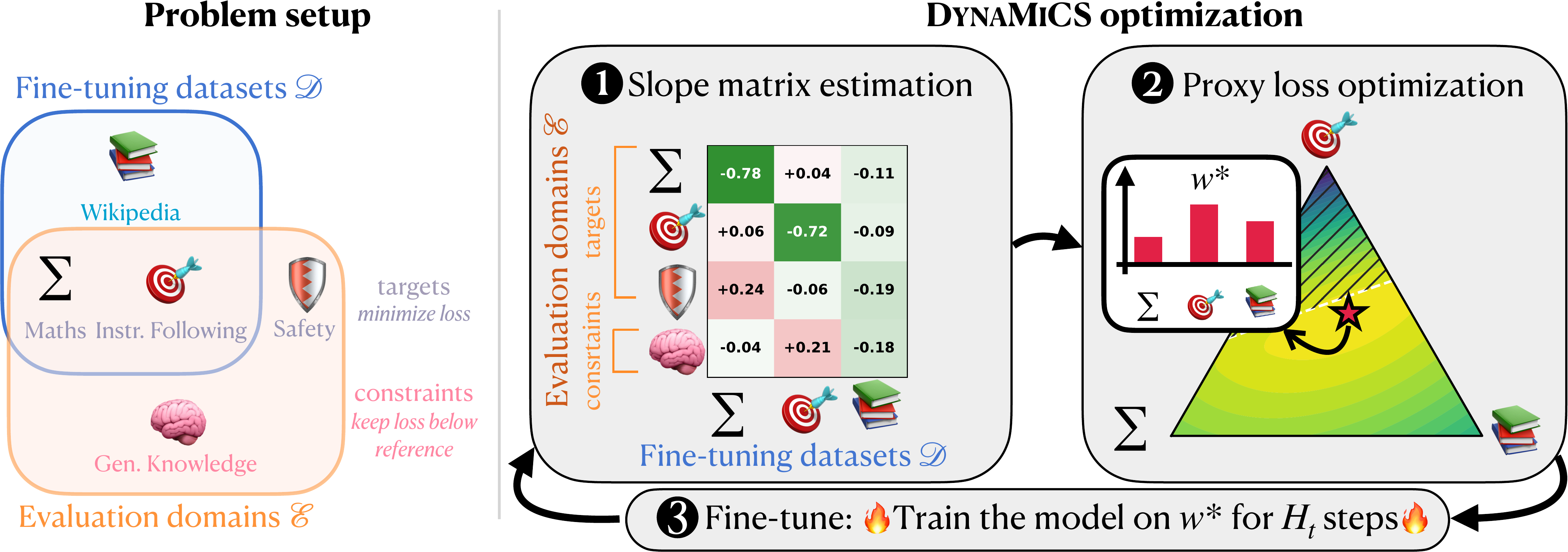} 
    \caption{
    \textbf{Problem setup.}
    Fine-tuning datasets $\mathcal{D}$ provide the data for mixture selection: target datasets plus optional auxiliary ones for transfer or regularization.
    Evaluation domains $\mathcal{E}$ are partitioned into target domains, whose losses are minimized, and constrained domains, whose losses must remain below reference values.
    \textbf{\method optimization.}
    At each update, \method estimates a slope matrix $\mathbf{S}(t)$ \textbf{(1)}, where $S_{ij}(t)$ measures the local effect of $D_j$ on evaluation loss $L_i$.
    Green/red entries denote loss decrease/increase.
    Given $\mathbf{S}(t)$, \method solves a constrained optimization problem to obtain weights $\mathbf{w}^*$ \textbf{(2)}, trains with them for $H_t$ steps \textbf{(3)}, and then repeats.
    The simplex illustrates the proxy objective landscape; white lines mark constraint boundaries; values are illustrative.\looseness-1
    \vspace*{-0.2cm}
    }

    \label{fig:teaser} 
\end{figure}

\textbf{Controlled Dynamic Mixtures}  
We propose \textit{dynamic mixtures for constraint satisfaction} (\method), a dynamic mixture optimizer that estimates local cross-domain effects and uses them to compute constraint-aware sampling weights during training, see Figure~\ref{fig:teaser}. We consider three types of knowledge domains: \emph{fine-tuning} datasets, from which the mixture is computed; and two \textit{evaluation} domains: \emph{target} domains, on which performance should be improved; and \emph{constrained} domains, for which the model's performance should remain above a reference level. \method periodically probes the impact on evaluation domains of fine-tuning for a few steps, using each of the fine-tuning datasets in isolation. The resulting loss changes across all evaluation domains form a \textit{slope matrix}, a "finite-differences" estimate of each fine-tuning dataset's effect on targets and constraints. Using this simple model, \method computes the optimal mixture weight that would achieve the best possible outcome (best improvement on target evaluations while remaining feasible), and commits to that mixture for a certain number of steps, before reevaluating the slope matrix. 
Our contributions are the following:
\begin{itemize}[nosep, leftmargin=*]
    \item \textbf{Constrained fine-tuning formulation.} We cast multi-domain fine-tuning as constrained optimization, where mixture weights are chosen to improve target domains while preserving performance on constrained capabilities.

    \item \textbf{Efficient constraint-aware mixture updates.} We propose \method, which converts short domain-specific probing runs into local estimates of transfer and interference, in each task's own units (perplexity or accuracy alike, since effects are measured by finite differences rather than gradients). Building on probing mechanisms used in prior work~\citep{skillit, aioli}, \method turns these local measurements into constraint-aware mixture updates that explicitly trade off improvement against capability preservation.\looseness-1

    \item \textbf{Comprehensive empirical analysis.} We evaluate \method on 50 multi-domain fine-tuning scenarios spanning 1--3 target domains and 3--10 constrained domains, testing Gemma3-12B~\citep{gemma3technicalreport}, Qwen3-8B~\citep{qwen3technicalreport}, and Qwen2.5-3B~\citep{qwen25}; the latter, for multiple LoRA ranks as well as full-parameter fine-tuning, with objectives and constraints expressed as losses or as benchmark accuracies. Across over 8,500 fine-tuning runs, and against fixed-weight sweeps and five mixture optimization methods, \method achieves reliable constraint satisfaction, which remains a challenge for all baselines, while reaching stronger target-domain improvements at substantially lower cost than best-of-many baseline selection.
\end{itemize}

%% file: sections/related_work.tex
\section{Related Work}
\label{sec:related-work}

\paragraph{Forgetting, overfitting, and data mixtures.}
Fine-tuning pre-trained language models on specialized domains can degrade previously acquired capabilities, a phenomenon known as \textit{catastrophic forgetting}~\citep{mccloskey:catastrophic, shi2024continuallearninglargelanguage, Kirkpatrick_2017}. In SFT, this risk is compounded by the small size of specialization datasets relative to pre-training corpora: models may adapt rapidly to the target domain while overfitting it and losing broader generalization ability~\citep{shi2024instructiontuninglossinstructions}. A common mitigation is to fine-tune on \emph{data mixtures}, interleaving target-domain data with supporting or general-domain data that can act as regularizers~\citep{bethune2025pretraining_injection, he2021analyzingforgettingproblempretrainfinetuning,DBLP:journals/tmlr/IbrahimTGRABLR24}. This raises the central question of how to choose mixture weights to improve target domains while limiting regressions on domains that should remain stable.

\paragraph{Static and dynamic mixture selection.}
Standard mixture choices include uniform sampling, sampling proportional to dataset size, and temperature-based sampling~\citep{wang2020balancingtrainingmultilingualneural, mueller-etal-2024-multi}. More principled static approaches formulate mixture selection as an optimization problem, for example, using scaling-law predictions to estimate useful mixture weights~\citep{DBLP:conf/icml/Li0X25}. However, static mixtures cannot react to evolving transfer and interference.
Recent approaches adapt the mixture online, using staged curricula~\citep{dong2024abilitieslargelanguagemodels}, reinforcement-learning controllers~\citep{wu2024mixtureofskillslearningoptimizedata, DBLP:journals/corr/abs-2508-12116}, loss-based signals~\citep{lu2025versatune}, sample-level influence estimates~\citep{EVIC}, or explicit models of cross-domain effects~\citep{skillit, aioli, tandem} (Appendix~\ref{app:related-work}). Among the latter, Skill-It and Aioli are closest to \method: Skill-It builds a skill-influence matrix, a structural precursor of our slope matrix, though non-negative and estimated once, while our local predictor has the same linear form as Aioli's dynamic mixing law. Across these methods, however, mixture selection remains unconstrained: even those that explicitly account for forgetting~\citep{lu2025versatune} offer no way to name a capability to preserve and bound its degradation.

\paragraph{Relation to pre-training mixture optimization.}
Mixture optimization has also been studied in pre-training, including proxy-model methods such as DoReMi~\citep{doremi} and DoGE~\citep{doge}, benchmark-driven data curation~\citep{li2024datacomp}, scaling-law-based mixture prediction~\citep{yedata}, online reweighting~\citep{zhaorethinking}, gradient-alignment methods~\citep{fan2024dynamicgradientalignmentonline}, and multi-target or domain-affinity-based reweighting frameworks~\citep{fangrape, xie2025chameleonflexibledatamixingframework}. DoReMi minimizes the worst-case excess loss across domains with Group-DRO on a small proxy model, and transfers the time-average of the resulting weights to the larger model as a single static mixture; DoGE instead combines independently estimated per-domain effects. In both cases the mixture is fixed before the final run: relative to a sweep over fixed weights, such methods contribute one additional candidate mixture, chosen without any notion of feasibility. Moreover, our setting differs from pre-training mixture optimization: in SFT, data is smaller, repeated more often, and forgetting of specific capabilities, such as alignment, is often a central concern. An extended discussion of this body of work is provided in
Appendix~\ref{app:related-work}.

%% file: sections/our_method.tex
\section{\method: Dynamic, Slope-Informed Data Mixture Optimization}
\label{sec:method}
We consider a fine-tuning setting where data from $N$ \emph{fine-tuning datasets} $\mathcal{D} = \{D_1, \ldots, D_N\}$ can be accessed, and each of $M$ \emph{evaluation domains} can be evaluated through a loss (or negative performance metric) $L_i, i\leq M$, where $\{1,\dots,M\}$ is split into two disjoint target and constrained sets of indices, $\mathcal{T}\cup\mathcal{C}= \{1,\ldots,M\}$.
The fine-tuning datasets and evaluation domains may overlap %
but need not coincide: target or constrained domains may be benchmarks not included in training, such as safety or general-knowledge evaluations, while other domains may enter the mixture without being optimization targets.
This includes settings in which a practitioner has one or a few small specialized datasets they want the model to learn, auxiliary corpora that may stabilize training, or held-out benchmarks on which performance should be preserved (e.g., safety or commonsense knowledge observable via a handful of evaluation batches, but for which training data may not be available).\looseness-1

\subsection{Constrained data-mixture optimization}
\label{sec:formulation}
Training proceeds on a mixture of fine-tuning datasets with weights $\mathbf{w} = (w_1, \ldots, w_N) \in \Delta^{N}$, where $\Delta^{N} = \{\mathbf{w} \in \mathbb{R}^N_{\geq 0} : \sum_j w_j = 1\}$ is the probability simplex.
At each training step $t$, a batch is drawn from fine-tuning dataset $D_j$ with proportion $w_j$. We formulate mixture selection as the constrained problem
\begin{equation}
\label{eq:main_problem}
\min_{\mathbf{w} \in \Delta^{N}}
\;\sum_{i \in \mathcal{T}} \hat{L}_i(\mathbf{w})
\quad \text{s.t.} \quad
\hat{L}_i(\mathbf{w})  \leq L_i^{\text{ref}},
\quad \forall\, i \in \mathcal{C},
\end{equation}
where $\hat{L}_i(\mathbf{w})$ is the \textit{predicted} loss on the $i$-th evaluation domain after training with weights $\mathbf{w}$ for a fixed horizon, and $L_i^{\text{ref}}$ is the reference loss,  computed on the base model prior to fine-tuning, for constrained domain $E_i$.
The central challenge is estimating $\hat{L}_i(\mathbf{w})$ reliably; we address this by measuring the marginal effect of each fine-tuning dataset independently and combining these effects through a local linear approximation.

\subsection{Objective and constraint approximation using finite-differences}
\label{sec:slopes}

To approximate $\hat{L}_i$ used in \eqref{eq:main_problem}, we resort to a ``finite differences'' approach to compute a linear approximation. We form a \emph{slope matrix} 
$\mathbf{S}(t) \in \mathbb{R}^{M \times N}$:
Each entry $S_{ij}(t)$ captures the change in the metric for evaluation domain $E_i$ induced by training on fine-tuning dataset $D_j$ exclusively for one step. 
An important flexibility provided by this approach is that computing the slope matrix does \textit{not} require evaluation domains (targets or constraints) to be represented within the fine-tuning datasets.

For each fine-tuning dataset $D_j$, starting from the current checkpoint (parameters and optimizer state), we perform $c_t$ temporary probing steps using only data from $D_j$ with the same optimizer used for the ongoing fine-tuning training. Both model and optimizer are restored after each domain's probe, so these updates serve only to estimate local loss changes. 

\begin{wrapfigure}{r}{.5\textwidth}
\begin{equation}
\label{eq:slope}
S_{ij}(t) = 
\frac{
\overbrace{L_{i}(\theta_j^{(c_t)})}^{\text{loss after probing}} 
- 
\overbrace{L_{i}(\theta_t)}^{\text{current loss}}
}{c_t}.
\end{equation}
\vskip-12pt
\end{wrapfigure}
We then measure the impact of this domain-specific probe on each evaluation domain:
Negative entries indicate that probing on $D_j$ decreases the loss on $E_i$, corresponding to beneficial transfer. Positive entries indicate an increase in loss, corresponding to interference or, when $E_i$ is a constrained domain, potential forgetting.

\subsection{Adaptive weight optimization}
\label{sec:optimization}

\method leverages the slope matrix to predict the effect of candidate mixtures on the evaluation losses.
Given $\mathbf{S}(t)$, we approximate the loss on evaluation domain $E_i$ after training for $H_t$ steps (the interval until the next weight update), which may vary across training following a schedule, with mixture weights $\mathbf{w}$ as in~\eqref{eq:prediction}. Equation~\ref{eq:prediction} is a local linear
\begin{wrapfigure}{r}{.5\textwidth}
\vskip-12pt
\begin{equation}
\label{eq:prediction}
\hat{L}_i(\mathbf{w}) =
\underbrace{L_{i}(\theta_t)}_{\text{current loss}}
+
H_t
\underbrace{\sum_{j=1}^{N} S_{ij}(t) w_j}_{\text{predicted change}}.
\end{equation}
\vskip-12pt
\end{wrapfigure}
approximation of the loss: per-domain effects are assumed to combine additively, and loss changes are extrapolated only over the short horizon between two mixture updates $H_t$.
Frequent re-estimation of $\mathbf{S}(t)$ ensures the reliability of this approximation, and we verify its fidelity directly (Appendix \ref{app:fidelity}). Note that we also explored a variant of this approach where we fit parametric curves to the loss trajectories observed during slope estimation, allowing for non-linear loss dynamics over the prediction horizon. This approach yields more accurate predictions but requires frequent $L_i$ evaluations to fit such curves. This inflates the overall compute budget, without significant performance gains (Appendix \ref{app:curves}).

Substituting~\eqref{eq:prediction} into~\eqref{eq:main_problem} and dropping terms constant in $\mathbf{w}$ yields a linear objective minimized over the simplex, with $\mathbf{S}_{i\cdot}(t)$ denoting the $i$-th row of $\mathbf{S}(t)$.  This optimization is a low-dimensional problem in $N$, the number of fine-tuning datasets.
Problem ~\eqref{eq:main_problem} is a linear program. However, we relax the hard constraints into a squared-hinge penalty so that the optimizer returns a least-violating mixture when no feasible point exists, a practically important regime:
\begin{equation}
\label{eq:penalized}
\min_{\mathbf{w} \in \Delta^N}\;
\underbrace{
\textstyle\sum_{i \in \mathcal{T}} \mathbf{S}_{i\cdot}(t)^\top \mathbf{w}
}_{\text{minimize target loss}}
\;+\;
\lambda
\underbrace{
\textstyle\sum_{i \in \mathcal{C}}
\max\!\Bigl(0,\;
\overbrace{\hat{L}_i(\mathbf{w}) - L_i^{\mathrm{ref}}}^{\text{constraint violation}}
+ \varepsilon \Bigr)^{\!2}
}_{\text{constraint penalties}},
\end{equation}
where $\hat{L}_i(\mathbf{w})$ is the predicted loss from~\eqref{eq:prediction} with prediction horizon $H_t$, $\lambda > 0$ controls the penalty strength, and $\varepsilon \geq 0$ is a safety margin that activates the penalty before the constraint boundary is reached.
At each update, we solve~\eqref{eq:penalized} over a grid of $(\lambda,\varepsilon)$ values and select the feasible solution with the lowest target objective, or the least-violating solution if none is feasible (Appendix \ref{app:training_hp}). The cost of these computations is negligible.
The resulting weights $\mathbf{w}^*$ are used as the mixture proportions until the next scheduled update (Figure~\ref{fig:teaser}). At designated training steps, \method re-estimates $\mathbf{S}(t)$ and recomputes $\mathbf{w}^*$.
The update schedule, defined as $\mathcal{U}$, is configurable: it can be a fixed interval, a geometric progression with doubling intervals, or any user-defined sequence. The full training loop is given in Algorithm~\ref{alg:dynamix}; Figure \ref{fig:teaser} offers an illustration of \method; hyperparameter choices ($\mathcal{U}$, $c$) and compute costs are detailed in Section~\ref{sec:setup}. 

\noindent
\begin{minipage}[t]{0.49\textwidth}
\vspace{0pt} %
\begin{algorithm}[H]
\small 
\caption{Slope matrix estimation}
\label{alg:slope_matrix}
\begin{algorithmic}[1]
\REQUIRE Current parameters $\theta_t$; optimizer state; fine-tuning datasets $\mathcal{D}$; evaluation domains $\mathcal{E}$; probing budget $c_t$
\ENSURE Slope matrix $\mathbf{S}(t) \in \mathbb{R}^{M \times N}$

\STATE \textbf{Evaluate current losses:} $L_i^{(0)} \leftarrow L_{i}(\theta_t)$ for all $E_i \in \mathcal{E}$

\FOR{each fine-tuning dataset $D_j \in \mathcal{D}$}
    \STATE Restore $\theta_t$ and optimizer state \hfill \textit{// temp probe}
    \STATE Fine-tune $c_t$ steps on $D_j$ only, yielding $\theta_j^{(c_t)}$

    \STATE \textbf{Fill column $j$ in $S$:}
    \STATE $S_{ij}(t) \leftarrow \dfrac{L_{i}(\theta_j^{(c_t)}) - L_i^{(0)}}{c_t} \quad \forall E_i \in \mathcal{E}$
\ENDFOR

\RETURN $\mathbf{S}(t)$
\end{algorithmic}
\end{algorithm}
\end{minipage}%
\hfill%
\begin{minipage}[t]{0.4925\textwidth}
\vspace{0pt} %
\begin{algorithm}[H]
\small 
\caption{Adaptive mixture optimization}
\label{alg:dynamix}
\begin{algorithmic}[1]
\REQUIRE Initial model $\theta_0$; fine-tuning datasets $\mathcal{D}$; eval domains $\mathcal{E}$; target set $\mathcal{T}$; constrained set $\mathcal{C}$; total steps $T_{\mathrm{total}}$; schedule $\mathcal{U} = \{0, t_1, \ldots\}$; budget $c$
\ENSURE Fine-tuned model

\STATE Evaluate reference losses: $L_i^{\text{ref}} \leftarrow L_{i}(\theta_0)$ $\forall$ $i \in \mathcal{C}$

\FOR{$t = 0, \ldots, T_{\mathrm{total}}-1$}
    \IF{$t \in \mathcal{U}$}
        \STATE $\mathbf{S}(t) \leftarrow \text{Alg.~\ref{alg:slope_matrix}}(\theta_t,\mathcal{D},\mathcal{E},c_t)$
        \STATE $\mathbf{w}^* \leftarrow \arg\min$ of~\eqref{eq:penalized} using $\mathbf{S}(t)$
    \ENDIF
    
    \STATE Train one step on mixture $(\mathcal{D},\mathbf{w}^*)$ to obtain $\theta_{t+1}$
\ENDFOR

\RETURN $\theta_{T_{\mathrm{total}}}$\hfill\textit{// post-hoc: select best}
\vspace{1.5em} %
\end{algorithmic}
\end{algorithm}
\end{minipage}

%% file: sections/experimental_setup.tex
\section{Experimental Setup}
\label{sec:setup}

\subsection{Evaluation scenarios}
We evaluate \method across 50 scenarios designed to cover increasingly challenging constrained fine-tuning settings. Each scenario defines a set of fine-tuning datasets, here split into target domains whose loss is explicitly minimized and non-target domains that provide additional data for mixture selection. 
It also defines constrained domains, which are used only for evaluation and whose loss must remain below the pre-fine-tuning reference. Targets span mathematical reasoning, code generation, general instruction following, medical QA, function calling, and multilingual tasks. Constrained domains represent capabilities to preserve, such as safety, commonsense reasoning, or general knowledge. Non-target domains range from generic corpora such as Wikipedia~\citep{wikipedia} to domain-specific datasets whose transfer relationship to the targets is not known a priori. The number of targets ranges from 1 to 3 and the number of constraints from 3 to 10. Scenario definitions and dataset descriptions are given in Appendices~\ref{app:datasets} and~\ref{app:scenarios}.

Since \method estimates domain effects through finite differences to actual fine-tuning steps rather than through gradients, its applicability is not restricted to differentiable objectives. This same property also lets a practitioner designate as a target a capability for which \emph{no training data is available}, since targets need not be represented among the fine-tuning datasets. We exploit both forms of flexibility in variants of the same 50 scenarios: promoting a held-out benchmark to target and scoring it by accuracy, which stresses the two at once since no training data optimizes it directly (Section~\ref{sec:results}); and expressing a constraint in accuracy rather than loss (Appendix~\ref{app:accuracy_experiments}). In both, a scenario mixes objectives, with loss used for some domains and accuracy for others.

\subsection{Models and training}
\label{sec:models}

We evaluate \method across three models, spanning 3B to 12B parameters. We fine-tune models in the following setups:
\begin{itemize}[nosep, leftmargin=*]
    \item \textbf{Qwen2.5-3B}~\citep{qwen25} with LoRA~\citep{lora} ranks $r \in \{8, 32, 64\}$, as well as full-parameter fine-tuning, to test sensitivity to the number of trainable parameters.
    \item \textbf{Qwen3-8B}~\citep{qwen3technicalreport} at $r\!=\!32$.
    \item \textbf{Gemma3-12B}~\citep{gemma3technicalreport} at $r\!=\!32$.
\end{itemize}
Note that \method is not tied to LoRA: all our optimization steps are equally applicable to full fine-tuning, and we verify this empirically on all 50 scenarios (Appendix \ref{app:full_FT}). We use LoRA as it provides a practical instantiation of fine-tuning, allowing for controlling adapter capacity.
We train all models with Adam for 2{,}048 steps with batch size 8, sequence length 512, and evaluations every 64 steps on 200 batches.
Training runs for a fixed budget without early stopping; the best feasible checkpoint is selected post-hoc (Section~\ref{sec:metrics}).
The learning rate is selected per model from a six-candidate sweep: $10^{-4}$ for Qwen2.5-3B (all ranks), $5\!\times\!10^{-5}$ for Qwen3-8B and Gemma3-12B, and $10^{-6}$ for full-parameter fine-tuning of Qwen2.5-3B. Further details on learning rate sweep, dataset sizes, token counts, and epoch behavior are provided in Appendix~\ref{app:training_hp}.

\subsection{\method configurations}
\label{sec:schedules}

Each of the model configurations is evaluated on all 50 scenarios with 3 random seeds.
In the main results, all optimizer hyper-parameters are held fixed; we vary only the update schedule $\mathcal{U}$, comparing denser versus sparser re-estimation of mixture weights in the early training phase. Our hypothesis is that early updates can help the optimizer gather information quickly when the loss landscape changes fastest; later, as the model stabilizes, less frequent updates suffice. Sensitivity to overall tightness of $\mathcal{U}$ and to the probe length is analyzed separately in Appendix~\ref{app:alpha_sweep}.\looseness-1

All three update at step~0 and share the same doubling tail  ---updates at steps 64, 128, 256, 512, 1024--- but differ in how aggressively they front-load early updates:
\begin{itemize}[nosep, leftmargin=18pt] 
    \item \emph{No warmup}: no additional updates (6 updates total).
    \item \emph{Light warmup}: additional updates at steps 8, 16, 32 (9 updates total).
    \item \emph{Dense warmup}: additional updates at steps 2, 4, 8, 16, 32 (11 updates total).
\end{itemize}
At each update step $t$, the number of probing steps $c_t$ used to estimate the slope matrix is set to $\min(H_t, c_{\max})$, with $c_{\max}\!=\!128$. In other words, early updates with short intervals use short probes (e.g.\ $c_t = 2$ at the first dense-warmup update), while later updates use a full budget of 128 steps.

\subsection{Baselines}
\label{sec:baselines}

\paragraph{Fixed-weight grid search.}
Our primary baseline holds mixture weights fixed for the entire run.
A practitioner facing this problem must choose a weight allocation without knowing how each dataset will interact with the targets and constraints, so a competitive fixed mixture can require an expensive sweep over candidate target/additional-domain trade-offs, a well-known bottleneck in data mixture optimization~\citep{doremi,yedata, kendall2018multitasklearningusinguncertainty}.
We represent this class of strategies with two allocation schemes that cover the most natural choices.
\emph{Uniform} splits a target mass $w$ equally across all target domains in the fine-tuning set, and the remaining $1 - w$ equally across all non-target domains in the fine-tuning set. 
\emph{Proportional}, instead, distributes $w$ across target datasets in proportion to their sample counts (giving more weight to larger datasets), while still splitting $1 - w$ uniformly across non-target datasets.
We sweep $w \in \{0.0, 0.2, 0.5, 0.8, 1.0\}$ and run both schemes, yielding up to 9 baseline runs per scenario. Proportional runs are skipped when they coincide with uniform, e.g., at $w = 0$ where all weight goes to non-target domains regardless of scheme. %
We report results of this baseline throughout the paper.\looseness-1

\paragraph{Additional baselines.}
We further compare \method against DoReMi~\citep{doremi}, the probing-based methods Skill-It~\citep{skillit} and Aioli~\citep{aioli}, a gradient-alignment baseline using instantaneous gradient dot products~\citep{ren2019learningreweightexamplesrobust,yu2020gradientsurgerymultitasklearning,fan2024dynamicgradientalignmentonline}, and an embedding-similarity variant that replaces probing with cosine similarities between domains.
None of these methods was originally designed to incorporate knowledge preservation constraints; we extend them to the constrained setting where possible. Overall, these methods fall short for one of two reasons: some rely on signals that are not calibrated in loss units and so cannot anticipate where a constraint will land, while others optimize an aggregate objective that does not distinguish target improvement from knowledge preservation, leaving them no way to trade one against the other.
Details and results are in Appendix~\ref{app:baselines}.

\subsection{Compute cost}
\label{sec:compute_cost}

\begin{wrapfigure}{r}{.45\textwidth}
\vskip-12pt
\begin{equation}
\label{eq:cost_ratio}
\beta_t =
\frac{N \times (\,\overbrace{c_t}^{\text{probing}} + \underbrace{C_{\text{eval,red}}}_{\text{reduced eval}}\,)}{\rho_t}
\end{equation}
\vskip-12pt
\end{wrapfigure}
We express all costs as multiples of a single baseline run ($1\times$), so that \method's overhead can be compared directly to the cost of running additional fixed-weight trials.
\method shares the training loop of the baselines but adds a per-update overhead: at each scheduled update step $t \in \mathcal{U}$, it probes each of the $N$ fine-tuning datasets in isolation.
This involves $c_t$ probing steps per dataset, followed by a reduced evaluation (in our experiments, 50 batches per domain vs.\ 200 in the main evaluations) on the evaluation domains.
The overhead thus scales with the number of candidate fine-tuning sources, not with the number of targets or constraints: an additional evaluation domain adds a few evaluation batches after each probe, but no further probing.
Pre-probe losses need not be recomputed either: they are recycled from the main evaluation loop whenever an update aligns with a scheduled evaluation.

For an update interval of length $H_t$ with a baseline compute cost of $\rho_t$, the relative per-update overhead ratio $\beta_t$ is given by~\eqref{eq:cost_ratio}, where $C_{\text{eval,red}}$ is the cost of one reduced evaluation.
Under the geometric schedule (Section~\ref{sec:schedules}), the adaptive rule $c_t = \min(H_t, c_{\max})$ ties probing to interval length: early short-interval updates use few probing steps, while later long intervals make the capped $c_{\max} = 128$ negligible relative to $\rho_t$.

We exclude the cost of solving ~\eqref{eq:penalized}: a low-dimensional simplex optimization on CPU, negligible next to GPU training and evaluation. Consequently, while a small grid search of 9 independent mixtures costs $9\times$ a single baseline run, \method totals only $1.4\times$ to $1.7\times$. These are means over all 50 scenarios, which span 1--3 target and 3--10 constrained domains, hence up to 13 evaluation domains. The detailed cost derivation, a per-schedule breakdown, and projected costs for larger numbers of sources and evaluation domains are provided in Appendix~\ref{app:compute_costs}. \looseness-1

\subsection{Metrics}
\label{sec:metrics}

\paragraph{Feasibility.}
Every evaluation domain is scored by a quantity for which lower is better: a loss, or, for benchmark-based domains, an error rate ($1 - \text{accuracy}$); a single scenario may mix the two. A run is \emph{feasible} if it produces at least one checkpoint where the score on \textit{every} constrained domain remains at or below its reference $L_i^{\text{ref}}$ and the sum of target scores has decreased relative to step~0. This means that, effectively, fine-tuning led to learning without forgetting, and that a run achieving feasibility by suppressing target learning would score close to zero under the metric defined below. We study this trade-off directly in Appendix~\ref{app:unconstrained}.

\paragraph{Constrained performance improvement.}
Our primary metric is the improvement on the target domains, at the best feasible checkpoint, defined as the one minimizing evaluation-set target scores while satisfying all constraints. For loss-based targets this is the relative perplexity reduction (geometric mean over multiple targets) reported on a held-out test set to prevent overfitting to the selection criterion. Targets may also be scored by accuracy, in which case we report the gain on the target benchmark in percentage points.
Infeasible runs contribute no improvement (0\%): if no checkpoint satisfies the constraints, the best a practitioner can do is keep the pre-fine-tuning model. %

\textbf{Baselines: best of $k$.} To place baselines on the same cost axis, we ask: for a given budget of $k$ fixed-weight runs (each costing $1\times$), how much improvement (perplexity reduction or accuracy gain) can a practitioner expect from the best of those $k$ attempts? 
For each scenario, we estimate the expected best-of-$k$ by sampling without replacement from the scenario's baseline pool, then report the mean across scenarios.\looseness-1

%% file: sections/results.tex
\section{Results}
\label{sec:results}

\begin{figure}[t]
    \centering
    \includegraphics[width=\textwidth]{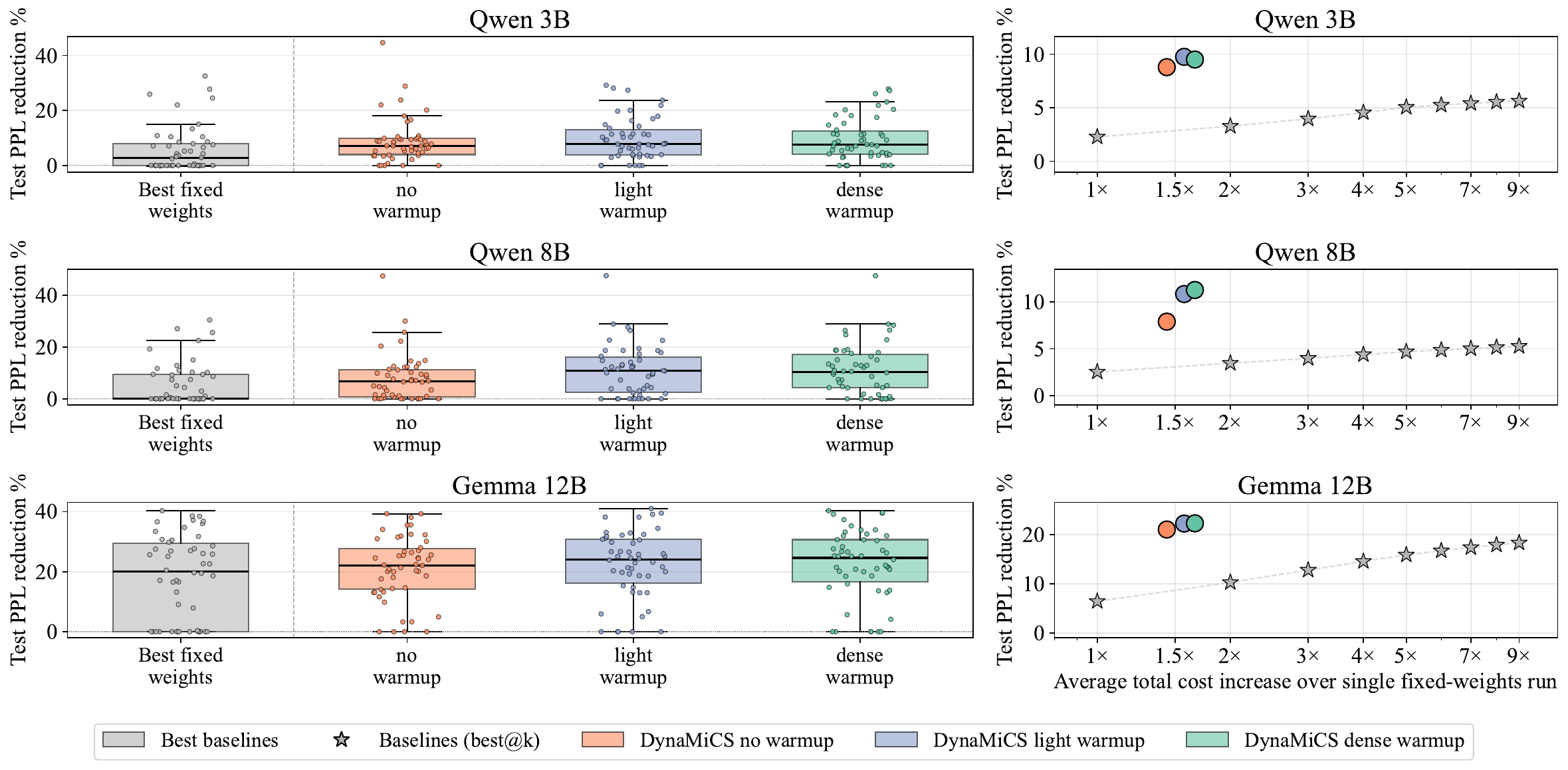}
    \vspace*{-0.2cm}
    \caption{\textbf{Model comparison.} Test perplexity reduction ($\%$) across 50 scenarios for different model families (Qwen2.5-3B, Qwen3-8B, Gemma3-12B), averaged over three seeds. Scenarios with no feasible checkpoint count as $0\%$. \textbf{Left (boxplots):} Per-scenario PPL reduction, one dot per scenario; for the baselines we report the oracle best across all fixed-weight configurations. \textbf{Right (scatter plots):} Mean PPL reduction across scenarios vs.\ compute cost (relative to a single fixed-weights run). Each \method colored circle corresponds to one warmup schedule and shows the mean reduction of a single adaptive run at that schedule's cost. Baseline stars show $best@k$: the best mean reduction obtained by \textit{k} fixed-weight configurations, at the corresponding budget. Higher is better.}

    \label{fig:model_comparison}
\end{figure}

We report here the results, across all 50 scenarios, of \method runs under the three geometric schedules described in Section~\ref{sec:schedules} comparing against fixed-weight baselines, for 3 random seeds (Figure~\ref{fig:model_comparison}).
Recall that infeasible runs, those where no fine-tuning checkpoint satisfies all constraints, contribute 0\% perplexity reduction, so the distributions reflect both the quality of feasible solutions and how often each method finds one. Feasibility rates are reported separately in Appendix~\ref{app:feasibility}.

\paragraph{\method consistently outperforms baselines.} Across all model configurations, \method with \textit{any} schedule achieves substantially higher mean test perplexity reduction than fixed-weight baselines. On Qwen2.5-3B, the light and dense warmup schedules reach a mean reduction of 9.5--9.8\% at ${\sim}1.7\times$ the cost of a single run; at a comparable budget of two fixed-weight attempts, baselines reach 3.3\%. %
On Qwen3-8B, the gap widens: \method achieves 11.3\% with dense warmup, against 3.5\% at two attempts. %
Even on Gemma3-12B, where baselines are considerably stronger, \method adds a further 4 percentage points over the entire grid of fixed-weight runs (22.3\% versus 18.3\%). Appendix~\ref{app:baselines} compares \method against five alternative mixture-selection methods on the same 50 scenarios (gradient alignment, embedding similarity, DoReMi, Skill-It and Aioli). None reaches a mean test perplexity reduction above $3.5\%$, with low feasibility rates; see Table~\ref{tab:baselines_summary}.\looseness-1

\begin{wrapfigure}{r}{0.55\textwidth}
    \vspace{-\baselineskip}   %
    \centering
    \includegraphics[width=\linewidth]{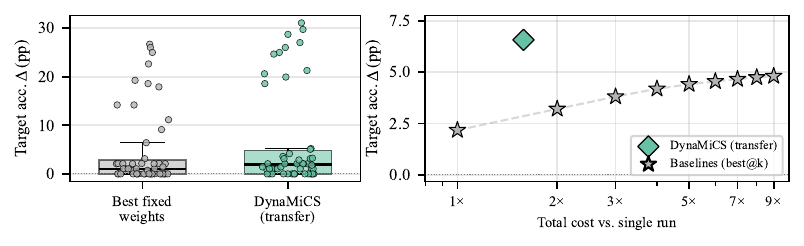}
     \captionsetup{skip=3pt}
      \caption{\textbf{Accuracy target on an unseen benchmark (Qwen2.5-3B, r=32).}
      Change in target-benchmark accuracy (pp) across the 50 scenarios.  Format as in Fig.~\ref{fig:model_comparison}. Higher is better.}
    \label{fig:accuracy_transfer_qwen}
    \vspace{-\baselineskip}   
\end{wrapfigure}
\paragraph{Beyond perplexity: non-differentiable, data-free targets.}
Because \method estimates domain effects by finite differences rather than gradients, it benefits from a flexibility that gradient-based mixture optimizers structurally lack: it can steer fine-tuning toward
\textit{any} target signal. We demonstrate this with an experiment that stresses \emph{two} forms of this flexibility at once: (i)~the target is scored by \emph{accuracy}, a non-differentiable metric; and (ii)~the target is \emph{unseen}, i.e.\ never part of the fine-tuning mixture, so no training data directly optimizes it. We rerun all 50 scenarios with Qwen2.5-3B (LoRA rank~32), designating in each scenario one held-out benchmark as an accuracy target (measured as the length-normalized log-likelihood of the choice text). Even in this harder setup, a single \method run raises held-out accuracy more than the best fixed-weight configuration (Figure~\ref{fig:accuracy_transfer_qwen}). The advantage holds for each target benchmark (see Appendix~\ref{app:accuracy_experiments}, which also covers accuracy used as a \emph{constraint}).

\paragraph{Results generalize across ranks, architectures, scales, and fine-tuning regimes.}
Varying the LoRA rank on Qwen2.5-3B from 8 to 64 has minimal effect on mean PPL reduction (8.7--9.8\% with warmup schedules).
An extended sweep over ranks 2--128 confirms that rank 32 sits at the knee of the capacity curve: lower ranks underperform while higher ranks offer no added benefit. \method also transfers to full-parameter fine-tuning (Appendix~\ref{app:rank_sensitivity} and~\ref{app:full_FT}).

\paragraph{Early updates matter.}
Warmup schedules consistently improve perplexity reduction over the no-warmup schedule, most clearly on Qwen3-8B (7.9\% to 10.8\% with light warmup). The difference between light and dense warmup is small in all configurations, suggesting that a moderate number of early updates suffices. On Qwen2.5-3B, denser warmup improves target quality but slightly reduces the feasibility rate (Appendix~\ref{app:feasibility}; see Appendix~\ref{app:alpha_sweep} for sensitivity to the overall update frequency).
\looseness-1

\begin{wrapfigure}{r}{0.4\textwidth}
\centering
    \vspace*{-0.5cm}
    \includegraphics[width=\linewidth]{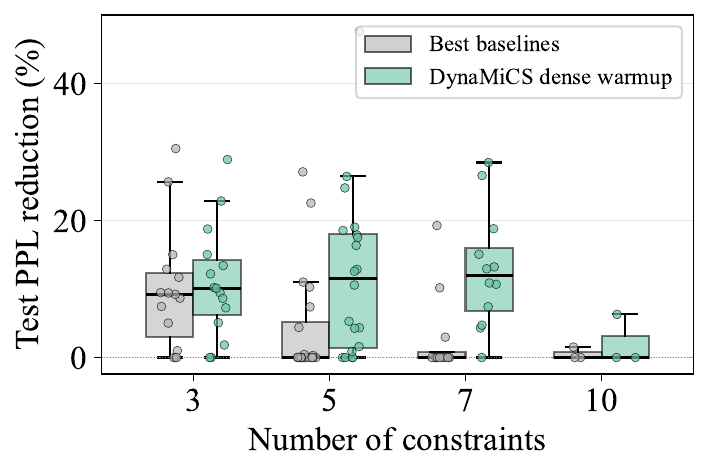}
    \vspace*{-0.6cm}
    \caption{Test PPL reduction by number of constraints (Qwen3-8B,
   LoRA-32). \method maintains gains as difficulty increases, while baselines degrade.\looseness-1}
   \vspace*{-0.75cm}
   \label{fig:by_constraints}
\end{wrapfigure} 
\paragraph{\method degrades gracefully.} %
As constraints grow, \method remains largely stable while baselines degrade steadily (Figure~\ref{fig:by_constraints}); with 10 constraints, both methods struggle, although \method still performs better for Gemma-3-12B and Qwen3-8B. Even when feasibility is not reached, \method yields lower constraint violations than the best baseline, showing that constraint-aware weighting limits forgetting even when full satisfaction is out of reach (Appendix~\ref{app:by_constraints} and~\ref{app:violation}).

\paragraph{Ablating dynamicity.}
The mixtures \method selects evolve differently across scenarios and models. A frequent pattern emphasizes specialized data early and shifts toward generic corpora later, where they act as regularizers (Appendix~\ref{app:weight}). To isolate what dynamic weight allocation contributes, we ablate it: we solve the constrained problem once at step~0 and freeze the resulting mixture, against the full schedule with dynamic updates (Qwen2.5-3B, $r{=}32$, all 50 scenarios, single seed). Feasibility is identical in both setups, while mean test-perplexity reduction rises from $2.2 \pm 3.6\%$ to $10.1 \pm 9.7\%$. The constrained formulation therefore delivers feasibility, and repeated re-optimization converts it into target improvement: a mixture chosen at step~0 preserves the protected capabilities but goes stale, whereas re-solving at each update keeps the path feasible while continuing to reduce the target loss.

%% file: sections/conclusion.tex
\section{Conclusion}

\method introduces an efficient framework for multi-domain fine-tuning by casting data mixture selection as a constrained optimization problem. By leveraging short domain-specific probing runs to estimate a slope matrix of cross-domain effects, it dynamically derives mixture weights that balance target improvement with explicit knowledge preservation constraints, without auxiliary models or manual tuning.
Because constraints enter through finite differences of evaluation metrics, any metric computable on a held-out batch can be used without a differentiable surrogate, a flexibility prior methods cannot offer by construction.
Across a wide range of scenarios, model families, and scales, this formulation yields consistent gains over baselines, achieving stronger target-domain performance while maintaining constraint satisfaction at a fraction of the cost of grid search.

\textbf{Limitations.} The method's reliance on local linear approximations can limit performance when update intervals are too long or when
feasible solutions become scarce in highly constrained settings. Additionally, the slope matrix assumes that cross-domain effects combine additively (a simplification shared with prior work). Moreover, the approach depends on the quality of evaluations, and may fail to preserve capabilities when these are small or unrepresentative. 
Finally, because mixture updates are chosen greedily from local estimates, early over-specialization may bias the subsequent training trajectory. Overcoming this locality typically requires back-tracking or evolutionary methods: an engineering challenge that we defer to future work.

%% file: sections/appendix.tex
\section{Extended Related Work}
\label{app:related-work}

\subsection{Forgetting and overfitting in LLM fine-tuning}

A fundamental challenge when fine-tuning a pre-trained language model on a specialized domain is \textit{catastrophic forgetting}: the model may lose capabilities acquired during pre-training as it adapts to the new distribution~\citep{mccloskey:catastrophic, shi2024continuallearninglargelanguage, Kirkpatrick_2017}. Its severity depends on factors including source--target similarity, gradient compatibility, and task ordering~\citep{kotha2024understandingcatastrophicforgettinglanguage, ramasesh2020anatomycatastrophicforgettinghidden, bell2022effecttaskorderingcontinual}.

This issue is especially relevant in SFT, where relatively small specialization datasets are often reused across epochs. Forgetting can then interact with \emph{overfitting}~\citep{shi2024instructiontuninglossinstructions}: the model may improve rapidly on the target domain while regressing on broader capabilities such as general knowledge, instruction following, or safety behavior. This motivates methods that improve target performance while controlling degradation on capabilities that should remain stable.

\subsection{Static data mixture selection}

A common way to mitigate forgetting and overfitting is to fine-tune on \emph{data mixtures}, interleaving specialization data with auxiliary or general-domain data~\citep{bethune2025pretraining_injection, he2021analyzingforgettingproblempretrainfinetuning, DBLP:journals/tmlr/IbrahimTGRABLR24}. The practical question is how to choose the mixture weights.
Standard choices include uniform, dataset-size-proportional, and temperature-based sampling~\citep{wang2020balancingtrainingmultilingualneural, mueller-etal-2024-multi}. These methods are simple and inexpensive, but their weights are fixed before training and cannot react to evolving transfer and interference. More principled static methods formulate mixture selection as an optimization problem, for example using scaling laws to predict useful mixture weights~\citep{DBLP:conf/icml/Li0X25}. Such approaches improve over hand-designed heuristics, but generally optimize a target or aggregate objective rather than enforcing explicit performance bounds on capabilities to be preserved.

\subsection{Dynamic data mixture selection}

Domain relationships do not remain fixed throughout training: data that is useful early in fine-tuning may become less useful later, while auxiliary data may become increasingly important as the model specializes. Indeed, interactions between training samples can evolve over training~\citep{EVIC}, motivating dynamic mixture adaptation.

Recent methods use several mechanisms. \citet{dong2024abilitieslargelanguagemodels} propose a dual-stage SFT recipe that shifts from specialized toward more general data. \citet{wu2024mixtureofskillslearningoptimizedata} adapt sampling through reinforcement learning, while \citet{DBLP:journals/corr/abs-2508-12116} formulate selection as a non-stationary multi-armed bandit. \citet{lu2025versatune} update domain weights using loss-derived notions of learnability and forgetting.
At finer granularity, EVIC~\citep{EVIC} estimates evolving gradient-based \emph{influence} between samples and uses it to construct an adaptive curriculum. Our method instead operates over mixture components or domains, reducing the cost of the control problem while still adapting over training.\looseness-1

A complementary family explicitly estimates cross-domain effects and uses them to update mixture weights~\citep{skillit, aioli, tandem}. Because these are the closest methodological precursors of \method, we discuss them separately in Appendix~\ref{app:probe-based}.
The key distinction is that \method adapts the mixture under explicit preservation constraints. Existing dynamic methods use staged schedules, rewards, influence scores, loss-derived signals, or estimated cross-domain effects; some explicitly account for forgetting, but none lets a practitioner specify both a capability to preserve and an allowable tolerance for its degradation.

\subsection{Explicit cross-domain influence estimation}
\label{app:probe-based}

The closest precursors of \method explicitly estimate how training on one data group affects performance on others, then use these estimates to re-weight the mixture. Their mechanisms differ, but they share the general pipeline of influence estimation followed by mixture adaptation. We describe the three closest methods below and the baseline instantiations of the first two in Appendix~\ref{app:baselines}.

\paragraph{Skill-It.}
Skill-It~\citep{skillit} is a structural precursor of our slope matrix. It learns a skill-influence graph from auxiliary runs that train on individual skills and measure changes in validation loss on evaluation skills, then uses this graph together with current losses to update sampling proportions online.
Two differences are important. First, the graph represents beneficial prerequisite relationships rather than transfer and interference; so harmful interactions are not encoded as negative edges. Second, it is estimated once and then held fixed. Skill-It can target an individual skill, but does not impose preservation bounds on other evaluation skills.

\paragraph{Aioli.}
Aioli~\citep{aioli} casts dynamic data mixing as optimization under a parametric \emph{mixing law} relating mixture weights to per-domain losses. In its linear form, a matrix of cross-domain coefficients captures how training proportions affect validation losses.
Our predictor has the same linear structure, but the coefficients are estimated differently. Aioli interleaves near-one-hot mixtures during training and infers cross-domain effects from the resulting loss changes along a single trajectory. \method instead branches controlled domain-specific probes from a common checkpoint, isolating each dataset's local effect from the same model state. Aioli optimizes an unconstrained validation-loss objective (either averaged across evaluation groups or, in its OOD extension, on a single target group) whereas \method imposes explicit performance constraints on capabilities to be preserved.

\paragraph{TANDEM.}
TANDEM~\citep{tandem} formulates mixture selection as a bi-level problem and approximates it with a penalized single-level objective. It uses twin networks, i.e., a proxy and a reference model, to estimate domain utility before updating the mixture, and evaluates the framework in both pre-training and SFT.
Its penalty enforces approximate optimality of the inner training problem, rather than preservation of a specific capability, while the outer objective remains aggregate validation loss. Moreover, the learned mixture is used in a separate final training run, introducing auxiliary optimization overhead.

\paragraph{Positioning of \method.}
\method likewise estimates cross-domain effects, but does so with controlled probes from the current checkpoint and represents both beneficial transfer and degradation. Because the effects are measured by finite differences in the evaluation metric itself, constraints can be expressed in loss, perplexity, or non-differentiable metrics such as benchmark accuracy. These estimates are then used in an explicitly constrained optimization problem that trades target improvement against practitioner-specified preservation bounds.

\subsection{Pre-training mixture optimization}

Data mixture optimization has also been widely studied in \emph{pre-training}~\citep{DBLP:journals/corr/abs-2507-09404}. The setting differs from SFT: pre-training typically involves long runs over massive datasets, whereas SFT uses smaller datasets that may be repeated across epochs and starts from a model whose existing capabilities may need to be preserved.

DoReMi~\citep{doremi} learns pre-training domain proportions using a reference model and a smaller proxy model. The proxy weights are updated by exponentiated gradient on excess loss under a Group-DRO objective, and their time-average defines a static mixture for the final model. DoGE~\citep{doge} similarly learns weights with a proxy model, using a bi-level generalization objective based on gradient alignment. TANDEM~\citep{tandem}, discussed above, extends this line with twin networks and also evaluates SFT.

Other approaches model the relationship between composition and performance. Data Mixing Laws~\citep{yedata} fit functions linking mixture proportions to per-domain losses and combine them with scaling laws to predict useful mixtures; the same framework also studies mixtures intended to mitigate catastrophic forgetting in continual training. DataComp-LM~\citep{li2024datacomp} studies benchmark-driven data filtering and mixture construction, while ADAPT~\citep{zhaorethinking} performs online sample-level reweighting in both pre-training and instruction tuning.

Gradient-based approaches include Dynamic Gradient Alignment (DGA)~\citep{fan2024dynamicgradientalignmentonline}, which upweights domains whose gradients align with a target set. GRAPE~\citep{fangrape} instead performs multi-target reweighting through a minimax objective, while Chameleon~\citep{xie2025chameleonflexibledatamixingframework} derives domain weights from a learned embedding-space affinity matrix.

These methods provide increasingly flexible mechanisms for choosing data mixtures, but generally optimize target, aggregate, or predictive criteria rather than explicit practitioner-specified preservation bounds. \method instead focuses on SFT and periodically estimates local domain effects from the current checkpoint to update mixture weights under explicit constraints on capabilities to be preserved.\looseness-1

\clearpage
\section{List of Datasets}
\label{app:datasets} 

Our evaluation suite draws on datasets spanning several categories: instruction-tuning corpora used as target or non-target fine-tuning datasets (and, at times, as constraints as well); open-domain text corpora used only as non-target fine-tuning datasets; and lightweight benchmarks used exclusively as constraint domains.
Table~\ref{tab:datasets} summarizes each dataset, its role in our experiments, and the abbreviation used throughout the paper.

\begin{table}[h]
\centering
\footnotesize
\setlength{\tabcolsep}{3.5pt}
\renewcommand{\arraystretch}{1.08}

\begin{tabularx}{\linewidth}{@{}
    l
    l
    >{\raggedright\arraybackslash}X
    l
@{}}
\toprule
\textbf{Abbrev.} & \textbf{Dataset} & \textbf{Description} & \textbf{Source} \\
\midrule

\multicolumn{4}{@{}>{\raggedright\arraybackslash}p{\linewidth}@{}}{
\textit{Instruction-tuning datasets used for fine-tuning or, in some scenarios, as evaluation-only constrained domains}
} \\
\addlinespace[2pt]

MM  & MetaMathQA            & Math QA bootstrapped from formal problems              & \citet{metamathqa} \\
NM  & NuminaMath-CoT        & Math reasoning dataset with chain-of-thought           & \citet{numinamath} \\
GS  & GSM8K                 & Human-written math word problems                       & \citet{gsm8k} \\
CF  & CodeFeedback          & Filtered code instruction--response pairs              & \citet{codefeedback} \\
AI  & AutoIF                & Instruction-following data generated via self-instruct pipelines & \citet{autoif} \\
OT  & OpenThoughts          & Mixed-domain reasoning in math, science, code, and logic & \citet{openthoughts} \\
OO  & OpenOrca              & Instruction-tuning data distilled from GPT-4 outputs   & \citet{openorca} \\
Ay  & AyaDataset            & Multilingual instruction-following across diverse tasks & \citet{ayadataset} \\
MF  & MedicalFlashcards     & Medical QA in flashcard format                         & \citet{medicalflashcards} \\
GF  & GlaiveFunctionCalling & Tool-use conversations with structured function calling & \citet{glaivefc} \\
MS  & MegaScience           & Scientific QA spanning multiple disciplines            & \citet{megascience} \\
BS  & BillSum               & Summarization of U.S.\ legislative bills               & \citet{billsum} \\
RQ  & RepliQA               & QA conditioned on a reference document                 & \citet{repliqa} \\

\midrule
\multicolumn{4}{@{}>{\raggedright\arraybackslash}p{\linewidth}@{}}{
\textit{Open-domain text corpora used as non-target fine-tuning domains only, with unlimited samples}
} \\
\addlinespace[2pt]

WP  & Wikipedia             & Encyclopedic corpus for general knowledge              & \citet{wikipedia} \\
FW  & FineWeb               & Large-scale filtered web corpus                        & \citet{fineweb} \\

\midrule
\multicolumn{4}{@{}>{\raggedright\arraybackslash}p{\linewidth}@{}}{
\textit{Evaluation benchmarks used as constrained domains only, never included in the training mixture}
} \\
\addlinespace[2pt]

tM  & tinyMMLU              & Multi-domain knowledge benchmark                       & \citet{tinybenchmarks} \\
tT  & tinyTruthfulQA        & Evaluates resistance to false or misleading outputs    & \citet{tinybenchmarks} \\
tH  & tinyHellaswag         & Commonsense reasoning via sentence completion          & \citet{tinybenchmarks} \\
tA  & tinyAI2\_ARC          & Science reasoning benchmark with exam-style MCQs       & \citet{tinybenchmarks} \\
NS  & NemotronSafety        & Evaluates safe refusal and handling of harmful queries & \citet{nemotronsafety} \\
SQ  & SafetyQA              & QA benchmark targeting safety and policy compliance    & \citet{safetyqa} \\
BQ  & BBQ                   & Bias evaluation in QA across demographic attributes    & \citet{bbq} \\

\bottomrule
\end{tabularx}
\vspace{5pt}

\caption{Datasets used in our experiments. In these scenarios, fine-tuning domains are divided into target domains, whose loss is explicitly minimized, and non-target domains, which provide additional data for mixture selection. Instruction-tuning datasets may appear as target or non-target fine-tuning domains, and some are held out as evaluation-only constraint domains depending on the scenario. Open-domain corpora (WP, FW) are used exclusively as non-target fine-tuning domains with unlimited samples. Evaluation benchmarks appear only as constraint domains and are never included in the training mixture. Abbreviations are used in Table~\ref{tab:scenarios}.}
\label{tab:datasets}
\end{table}

\clearpage
\section{50 Evaluation Scenarios}
\label{app:scenarios}

We evaluate across 50 scenarios of increasing difficulty. Each scenario specifies:
\begin{itemize}[nosep, leftmargin=18pt]
    \item \textbf{Fine-tuning datasets}: available training data. These are divided into:
    \begin{itemize}
        \item \textbf{Targets}: datasets whose loss we aim to minimize. When applicable, the fixed sample budget is shown in parentheses.
        \item \textbf{Non-targets}: auxiliary fine-tuning datasets included in the training mixture, but not treated as optimization targets.
    \end{itemize}
    \item \textbf{Constraints}: evaluation-only domains whose loss must not exceed the pre-fine-tuning baseline. They are not included in the fine-tuning data.
\end{itemize}

The number of targets ranges from 1 to 3, while the number of constraints ranges from 3 to 10, yielding scenarios of increasing difficulty.

\providecommand{\tinybm}{\ensuremath{\times}\text{tiny}}

\definecolor{grpA}{RGB}{237,243,252}  %
\definecolor{grpB}{RGB}{245,245,245}  %

\begin{table*}[p]
\centering
\scriptsize
\setlength{\tabcolsep}{3.5pt}
\renewcommand{\arraystretch}{1.12}
\begin{tabularx}{\linewidth}{@{}r@{\hskip 4pt}c@{\hskip 4pt}c@{\hskip 6pt}
    >{\raggedright\arraybackslash\hsize=0.88\hsize}X@{\hskip 4pt}
    >{\raggedright\arraybackslash\hsize=1.12\hsize}X@{\hskip 4pt}
    >{\raggedright\arraybackslash}X@{}}
\toprule
\multirow{2}{*}{\textbf{\#}} & \multirow{2}{*}{$\boldsymbol{n_T}$} & \multirow{2}{*}{$\boldsymbol{n_C}$} & \multicolumn{2}{c}{\textbf{Fine-tuning datasets}} & \multirow{2}{*}{\shortstack[l]{\textbf{Constraints}\\\textbf{(eval only)}}} \\
\cmidrule(lr){4-5}
& & & \textbf{Targets} & \textbf{Non-targets} & \\
\midrule
\rowcolor{grpA} 1  & 1 & 3  & MM\,(2k) & WP, RQ\,(5k), OO\,(10k) & tA, tH, MS \\
\rowcolor{grpA} 2  & 1 & 3  & CF\,(500) & FW, OT\,(2k), Ay\,(5k) & tT, tH, NS \\
\rowcolor{grpA} 3  & 1 & 3  & OT\,(1k) & WP, Ay\,(300), GS\,(5k) & tT, tH, BQ \\
\rowcolor{grpA} 4  & 1 & 3  & RQ\,(10k) & FW, MM\,(300) & tA, tH, tM \\
\rowcolor{grpA} 5  & 1 & 3  & MM\,(800) & WP, Ay\,(500), CF\,(300), OO\,(30k) & tA, SQ, NS \\
\rowcolor{grpA} 6  & 1 & 3  & NM\,(2k) & WP, FW, BS\,(15k) & tT, tA, BQ \\
\rowcolor{grpB} 7  & 1 & 5  & OO\,(1.5k) & FW, NM\,(500), Ay\,(300), BS\,(10k) & tT, tA, NS, BQ, MS \\
\rowcolor{grpB} 8  & 1 & 5  & AI\,(1k) & WP, MS\,(500), OT\,(8k) & tT, tH, tA, SQ, BQ \\
\rowcolor{grpB} 9  & 1 & 5  & GF\,(300) & FW, MF\,(500), NM\,(800), OO\,(5k) & tT, tA, tM, SQ, BQ \\
\rowcolor{grpB} 10 & 1 & 5  & MF\,(500) & WP, NM\,(300), Ay\,(1k), BS\,(10k) & tT, tH, tA, SQ, NS \\
\midrule
\rowcolor{grpA} 11 & 2 & 3  & OO\,(3k), AI\,(1.5k) & FW, WP, RQ\,(800) & tA, tM, SQ \\
\rowcolor{grpA} 12 & 2 & 3  & GF\,(500), CF\,(500) & WP, RQ\,(10k), OT\,(5k) & tA, tT, NS \\
\rowcolor{grpA} 13 & 2 & 3  & CF\,(500), MM\,(500) & FW, OT\,(15k), Ay\,(5k) & tT, SQ, NS \\
\rowcolor{grpA} 14 & 2 & 3  & MF\,(500), OO\,(1k) & FW, Ay\,(2k), GS\,(5k) & tM, BQ, SQ \\
\rowcolor{grpA} 15 & 2 & 3  & NM\,(3k), GS\,(500) & WP, RQ\,(2k), OO\,(10k) & tH, tA, BQ \\
\rowcolor{grpB} 16 & 2 & 5  & GS\,(500), OO\,(800) & WP, FW, BS\,(10k), RQ\,(8k) & 4\tinybm, MS \\
\rowcolor{grpB} 17 & 2 & 5  & MM\,(500), MF\,(300) & FW, OT\,(2k), Ay\,(800) & 4\tinybm, SQ \\
\rowcolor{grpB} 18 & 2 & 5  & MM\,(800), OO\,(1.5k) & WP, Ay\,(300), RQ\,(2k), BS\,(5k) & tT, tH, tA, BQ, NS \\
\rowcolor{grpB} 19 & 2 & 5  & RQ\,(8k), MM\,(800) & WP, FW, NM\,(500) & 4\tinybm, NS \\
\rowcolor{grpB} 20 & 2 & 5  & CF\,(300), AI\,(500) & BS\,(3k), OT\,(2k), GS\,(5k) & 4\tinybm, NS \\
\rowcolor{grpB} 21 & 2 & 5  & NM\,(1k), GF\,(300) & WP, MF\,(500), Ay\,(800), OO\,(10k) & 4\tinybm, SQ \\
\rowcolor{grpB} 22 & 2 & 5  & MM\,(500), OO\,(1k) & WP, FW & 4\tinybm, NS \\
\rowcolor{grpA} 23 & 2 & 7  & NM\,(800), MF\,(300) & WP, FW, RQ\,(3k), Ay\,(500), OT\,(5k) & 4\tinybm, BQ, NS, SQ \\
\rowcolor{grpA} 24 & 2 & 7  & Ay\,(500), GF\,(500) & WP, FW, OT\,(2k), MS\,(800) & 4\tinybm, NS, SQ, BQ \\
\rowcolor{grpA} 25 & 2 & 7  & MM\,(500), OO\,(1k) & WP, FW, RQ\,(10k), BS\,(8k), Ay\,(5k) & 4\tinybm, MS, NS, BQ \\
\rowcolor{grpA} 26 & 2 & 7  & CF\,(200), AI\,(800) & WP, FW, OT\,(2k), Ay\,(500) & 4\tinybm, SQ, BQ, MS \\
\rowcolor{grpA} 27 & 2 & 7  & NM\,(1k), OO\,(1.5k) & WP, FW, RQ\,(3k), MF\,(300), GS\,(10k) & 4\tinybm, NS, SQ, BQ \\
\rowcolor{grpA} 28 & 2 & 7  & NM\,(1k), GF\,(300) & WP, FW & 4\tinybm, SQ, BQ, MS \\
\midrule
\rowcolor{grpA} 29 & 3 & 3  & NM\,(5k), GS\,(500), MM\,(2k) & WP, FW & tT, NS, BQ \\
\rowcolor{grpA} 30 & 3 & 3  & OO\,(5k), Ay\,(500), MF\,(500) & FW, BS\,(10k) & tM, tA, BQ \\
\rowcolor{grpA} 31 & 3 & 3  & MM\,(500), OO\,(3k), CF\,(500) & WP, RQ\,(10k) & tA, tH, SQ \\
\rowcolor{grpA} 32 & 3 & 3  & AI\,(3k), GF\,(500), NM\,(500) & FW, OT\,(8k), Ay\,(5k) & tT, NS, MS \\
\rowcolor{grpB} 33 & 3 & 5  & MM\,(800), OO\,(1.5k), NM\,(500) & WP, FW & 4\tinybm, BQ \\
\rowcolor{grpB} 34 & 3 & 5  & MM\,(800), OO\,(1.5k), GF\,(300) & WP, BS\,(20k), Ay\,(50k) & 4\tinybm, MS \\
\rowcolor{grpB} 35 & 3 & 5  & AI\,(500), CF\,(300), NM\,(1k) & FW, OT\,(2k) & 4\tinybm, SQ \\
\rowcolor{grpB} 36 & 3 & 5  & OO\,(1k), MF\,(500), GS\,(500) & WP, Ay\,(3k), RQ\,(5k) & 4\tinybm, BQ \\
\rowcolor{grpB} 37 & 3 & 5  & MM\,(500), OO\,(1k), CF\,(300) & FW, RQ\,(10k), Ay\,(5k), BS\,(10k) & 4\tinybm, NS \\
\rowcolor{grpB} 38 & 3 & 5  & NM\,(800), MF\,(500), Ay\,(500) & WP, OT\,(1.5k), GS\,(5k) & tT, tA, BQ, NS, SQ \\
\rowcolor{grpB} 39 & 3 & 5  & GF\,(300), OO\,(1.5k), AI\,(500) & FW, BS\,(20k), RQ\,(10k) & 4\tinybm, MS \\
\rowcolor{grpB} 40 & 3 & 5  & OO\,(1k), MF\,(500), GS\,(500) & WP, FW & 4\tinybm, NS \\
\rowcolor{grpA} 41 & 3 & 6  & NM\,(800), AI\,(500), CF\,(300) & WP, FW & 4\tinybm, SQ, BQ \\
\rowcolor{grpB} 42 & 3 & 7  & AI\,(500), MF\,(300), GF\,(500) & WP, FW & 4\tinybm, NS, SQ, MS \\
\rowcolor{grpB} 43 & 3 & 7  & MM\,(800), OO\,(1.5k), CF\,(300) & WP, FW, RQ\,(2k), Ay\,(5k) & 4\tinybm, MS, SQ, BQ \\
\rowcolor{grpB} 44 & 3 & 7  & OO\,(1k), MF\,(300), AI\,(500) & WP, FW, RQ\,(3k), BS\,(10k) & 4\tinybm, BQ, NS, SQ \\
\rowcolor{grpB} 45 & 3 & 7  & NM\,(1k), Ay\,(300), GF\,(500) & WP, FW, OT\,(2k), RQ\,(1.5k) & 4\tinybm, MS, NS, BQ \\
\rowcolor{grpB} 46 & 3 & 7  & MM\,(500), OO\,(1.5k), CF\,(200) & WP, FW, OT\,(10k), BS\,(20k), Ay\,(5k) & 4\tinybm, NS, SQ, BQ \\
\rowcolor{grpB} 47 & 3 & 7  & AI\,(800), MF\,(300), GS\,(500) & WP, FW, Ay\,(500), RQ\,(3k), OO\,(10k) & 4\tinybm, BQ, NS, SQ \\
\rowcolor{grpA} 48 & 3 & 10 & MM\,(500), OO\,(1k), CF\,(300) & WP, FW, OT\,(10k), Ay\,(800), GF\,(5k) & 4\tinybm, MS, BS, RQ, NS, SQ, BQ \\
\rowcolor{grpA} 49 & 3 & 10 & Ay\,(500), GF\,(300), NM\,(1k) & WP, FW, OT\,(2k), RQ\,(3k) & 4\tinybm, MS, OO, BS, NS, SQ, BQ \\
\rowcolor{grpA} 50 & 3 & 10 & OO\,(1.5k), AI\,(800), CF\,(200) & WP, FW, MF\,(500), Ay\,(300), GS\,(5k) & 4\tinybm, MS, BS, RQ, NS, SQ, BQ \\
\bottomrule
\end{tabularx}
\caption{Our 50 evaluation scenarios. Sample budgets in parentheses. ``4\tinybm'' = tM, tT, tH, tA. \\WP and FW are regularizers with unlimited samples.}
\label{tab:scenarios}
\end{table*}

\clearpage
\section{Details on Hyper-parameters}
\label{app:training_hp}

\begin{figure*}[!ht]
\vspace*{-0.15cm}
    \centering
    \includegraphics[width=0.8\textwidth]{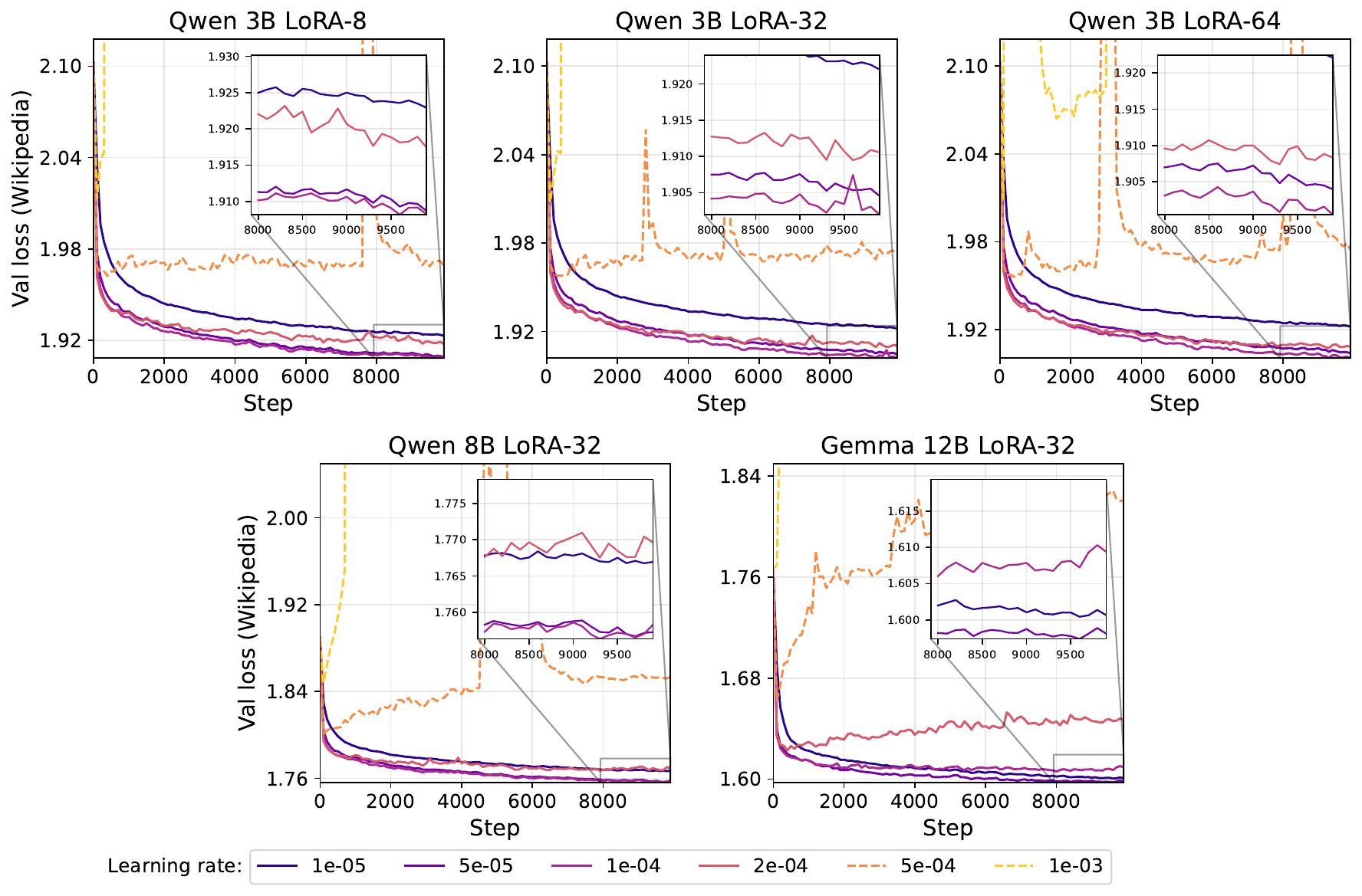}
    \caption{\textbf{Learning rate ablation.} Validation loss on Wikipedia across learning rates in Qwen2.5-3B at LoRA $r\!\in\!\{8,32,64\}$ (top) and Qwen3-8B and Gemma-3-12B at $r\!=\!32$ (bottom). Diverged runs are dashed and allowed to leave the cropped y-axis; insets zoom on the tail.}
    \label{fig:lr_sweeps}
\end{figure*}

\paragraph{Learning rate and warmup schedule.} We ran an exhaustive ablation over the learning rate and warmup schedule for fine-tuning on Wikipedia~\citep{wikipedia}. For each presented model and LoRA rank we swept six learning rates {1e-5, 5e-5, 1e-4, 2e-4, 5e-4, 1e-3}, as shown in Figure~\ref{fig:lr_sweeps}, crossed with three warmup configurations (0 steps, 1\%, and 2\% of total training steps). We found that warmup had no appreciable effect on the final validation loss in any setting, while the learning rates had a consistent optimum around 1e-4 for Qwen2.5-3B with ranks 8, 32, and 64, and 5e-5 for Qwen3-8B and Gemma3-12B, both with rank 32. Based on this, all results in the main text use no warmup, with the per-model best learning rate selected from this sweep.

\vspace*{-0.3cm}
\paragraph{Training budget.}
All runs train for 2{,}048 steps with batch size 8 and sequence length 512, corresponding to approximately 8$M$ training tokens per run.
We do not use early stopping: training proceeds for the full budget, and the best feasible checkpoint is selected post-hoc based on constraint satisfaction and target loss (Section~\ref{sec:metrics}).

\paragraph{Dataset sizes and epochs.}
Fine-tuning datasets vary in size from 300 to effectively unlimited examples (i.e., that will not be exhausted during the training phase), see Appendix \ref{app:scenarios}.
With mixture sampling, smaller datasets may be repeated multiple times (i.e., trained for more than one epoch), while larger datasets are subsampled.

\paragraph{Hyper-parameters of \method optimization.}

As mentioned in Section \ref{sec:method}, 
at each mixture update, we solve the proxy objective~\eqref{eq:penalized} over a two-dimensional grid of hyper-parameters $(\lambda, \varepsilon)$.
The penalty strength $\lambda$ is log-spaced over $[1, 5000]$ (15 values) and the constraint tolerance $\varepsilon$ is linearly spaced over $[0, 0.1]$ (3 values), yielding 45 candidate solutions per update.
Among candidates that satisfy all constraints, we select the one with the lowest target objective.
If no candidate is feasible, we select the one with the smallest maximum constraint violation.
This grid search is computationally cheap: all candidates are optimized over the same predicted loss surface (the fitted slope matrix), so the grid introduces no additional model evaluations beyond the single slope computation step.

\vspace*{-0.1cm}
\paragraph{Compute resources.}
\label{app:compute}

We trained over 8,500 models. Each run was performed on a single GPU node, using either H100 or B200 GPUs. For the bigger model (Gemma3-12B), wall-clock time for one \method run with dense warmup schedule, for s50, i.e., one of the largest sets of domains (Section \ref{sec:setup}), took approximately 10.5 hours. One fixed-weight baseline run for the same scenario has a wall-clock time of 7 hours.

\clearpage
\section{Compute Costs Analysis}
\label{app:compute_costs}

Both \method and the fixed-weight baselines share the same per-step training loop: a forward--backward pass on the current mixture, plus periodic evaluations (every 64 steps, 200 batches per dataset).
\method additionally runs \emph{slope computation} at each weight update: for every fine-tuning dataset, it trains briefly on that dataset alone ($c_t$ steps) and evaluates the result on all evaluation domains using a reduced budget of 50 batches per dataset.

\subsection{Linear weight update}
\label{app:linear_weight_update_costs}

The linear predictor requires two loss measurements per probe: one before (anchor) and one after $c_t$ training steps.
When a weight update aligns with a main evaluation step, the baseline losses are reused at no extra cost (\emph{evaluation recycling}); otherwise, one additional reduced evaluation is needed.
The per-update overhead relative to one baseline training interval $\rho_t$ is:
\begin{equation}
    \beta_t = \frac{N \times (c_t + C_{\text{eval,red}}) + \mathbf{1}_{\text{no recycle}} \cdot C_{\text{eval,red}}}{\rho_t},
\end{equation}
where $N$ is the number of fine-tuning datasets, $C_{\text{eval,red}} = M \times 50 \times C_{\text{fwd}}$ is the cost of one reduced evaluation over $M$ evaluation domains, $\rho_t = H_t \times C_{\text{bp}} + (H_t / 64) \times M \times 200 \times C_{\text{fwd}}$, $c_t = \min(H_t,\; c_{\max})$, $C_{\text{fwd}}$ the per-batch forward-pass cost, and $C_{\text{bp}}$ the cost of one training step on a batch. The total cost of a \method run is $\sum_{t \in \mathcal{U}}\rho_t(1 + \beta_t)$. The cost of solving Equation~\ref{eq:penalized} is negligible (a low-dimensional simplex problem on CPU) and excluded from the compute calculations.\looseness-1

\textbf{Cost of geometric schedules.} Table~\ref{tab:schedule_costs} shows the cost of each schedule discussed in Section~\ref{sec:results} averaged across all 50 scenarios (which vary in $N$ and $M$).
The overhead is low because the geometric schedules concentrate most training time in long later intervals where $\beta_t \ll 1$, and all updates in the geometric tail align with the evaluation frequency, enabling recycling.
Early warmup updates use short probes $c_t = \min(H_t, 128)$, so their absolute cost is small despite higher $\beta_t$.

\begin{table}[h]
\centering\small
\begin{tabular}{@{}lccc@{}}
\toprule
Schedule & Update steps & Updates & Mean cost \\
\midrule
No warmup & 0, 64, 128, 256, 512, 1024 & 6 & ${\sim}1.4\times$ \\
Light warmup & 0, 8, 16, 32, 64, \ldots, 1024 & 9 & ${\sim}1.5\times$ \\
Dense warmup & 0, 2, 4, \ldots, 32, 64, \ldots, 1024 & 11 & ${\sim}1.7\times$ \\
\bottomrule
\end{tabular}
\vspace{5pt}
\caption{Compute cost of the schedule variants discussed in Section \ref{sec:schedules} and their mean compute costs ($c_{\max}\!=\!128$, averaged across 50 scenarios).}
\label{tab:schedule_costs}
\end{table}

\textbf{Cost of fixed update horizons.} Results presented in Appendix~\ref{app:alpha_sweep} use a uniform update schedule with fixed interval $H_t \equiv H$ and $c_t \equiv H$ for all $t \in \mathcal{U}$, so that probing spans the full interval at every update.
Higher values of $H$ yield fewer total updates but higher per-update overhead.
Table~\ref{tab:alpha_costs} reports the mean cost for the three configurations discussed in the paper.

\begin{table}[h]
\centering\small
\begin{tabular}{@{}lcc@{}}
\toprule
Interval ($H = c_t$) & Updates & Mean cost \\
\midrule
128 & 16 & ${\sim}2.3\times$ \\
256 & 8 & ${\sim}2.0\times$ \\
512 & 4 & ${\sim}1.8\times$ \\
\bottomrule
\end{tabular}
\vspace{5pt}
\caption{Compute cost of schedules with fixed update horizons from Appendix \ref{app:alpha_sweep} (averaged across 50 scenarios).}
\label{tab:alpha_costs}
\end{table}

\textbf{Cost of non-linear weight update.} The curve-fitting predictor (Appendix~\ref{app:curves}) evaluates at $n_{\text{evals}} = 5$ intermediate checkpoints during each probe, rather than a single endpoint measurement.
This increases the evaluation cost per update by a factor of 5 relative to the linear predictor, while the probing training cost ($c_t$ steps) remains unchanged.
Table~\ref{tab:curves_costs} reports the cost for the same fixed-interval configurations with $H_t \equiv H$ and $c_t \equiv H$ for all $t \in \mathcal{U}$ presented in Table \ref{tab:alpha_costs}, when the curve-fitting predictor is used.

\begin{table}[H]
\centering\small
\begin{tabular}{@{}lcc@{}}
\toprule
Interval ($H = c_t$) & Updates & Mean cost \\
\midrule
128 & 16 & ${\sim}4.7\times$ \\
256 & 8 & ${\sim}3.2\times$ \\
512 & 4 & ${\sim}2.4\times$ \\
\bottomrule
\end{tabular}
\vspace{5pt}
\caption{Compute cost of the non-linear (curve-fitting) predictor with $n_{\text{evals}} = 5$ and fixed update horizons from Appendix~\ref{app:curves} (averaged across 50 scenarios).}
\label{tab:curves_costs}
\end{table}

\paragraph{Projected cost at larger problem sizes.}
The scenarios we run span $N \in [3,6]$ fine-tuning sources and up to 13 evaluation domains. To indicate how the overhead behaves beyond that range, Table~\ref{tab:cost_extrapolation} evaluates the cost model of Equation~\ref{eq:cost_ratio} at larger $N$ and $M$ under the dense warmup
schedule. Cost grows with the number of candidate sources, since each adds a probe, and is essentially flat across the columns, confirming that the overhead is driven by the number of sources rather than by the number of evaluation domains. Even the largest configuration remains inexpensive in absolute terms: a 10-target, 20-constraint problem with 12 candidate sources costs $2.15\times$, roughly the price of two fixed-weight runs for a problem with 30 competing objectives, and a budget of two fixed-weight runs yields very low feasibility rates (Figure~\ref{fig:model_comparison}).

\begin{table}[h]
  \centering
  \begin{tabular}{lccc}
  \toprule
  Fine-tuning sources $N$ & 3T / 10C & 5T / 15C & 10T / 20C \\
  \midrule
  5  & $1.54\times$ & $1.52\times$ & $1.50\times$ \\
  8  & $1.85\times$ & $1.81\times$ & $1.78\times$ \\
  12 & $2.25\times$ & $2.19\times$ & $2.15\times$ \\
  \bottomrule
  \end{tabular}
   \caption{Projected compute cost of \method relative to a single baseline run, obtained by evaluating Equation~\ref{eq:cost_ratio} at problem sizes larger than those we run (dense warmup schedule,
  $c_{\max}=128$). Columns give the number of target (T) and constrained (C) evaluation domains. Within a row, cost decreases slightly as
  evaluation domains are added because the $1\times$ reference grows faster than the overhead does: a baseline run evaluates every evaluation domain periodically at 200 batches, while the overhead adds
  only a single 50-batch reduced evaluation per probe, so the ratio of the two becomes smaller.}
  \label{tab:cost_extrapolation}
\end{table}

\clearpage
\section{Feasibility Rates}
\label{app:feasibility}

\begin{figure}[H]
    \centering
    \includegraphics[width=\textwidth]{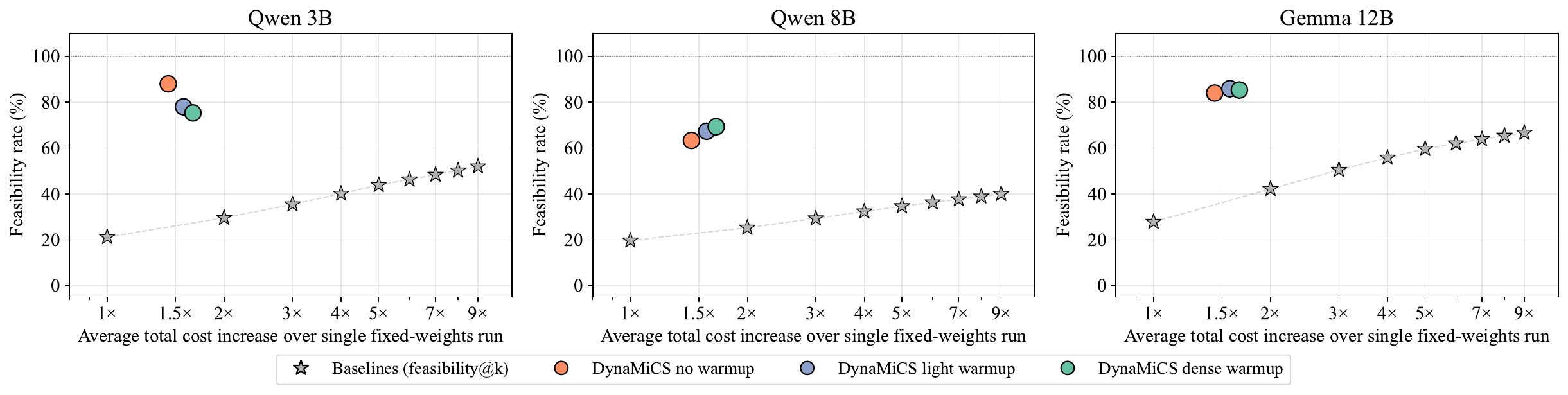}\\[0.5em]
    \includegraphics[width=\textwidth]{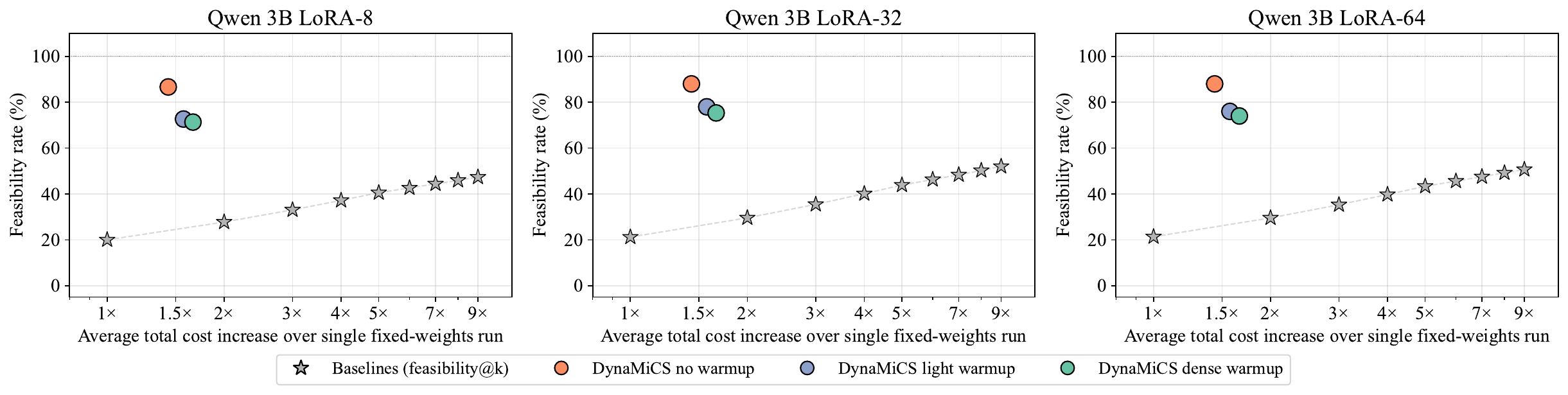}
    \caption{
    \textbf{Feasibility rate vs.\ compute cost} across 50 scenarios. 
    A run is counted as feasible only if a checkpoint \emph{after step~0} satisfies all constraints while improving target perplexity (the base model checkpoint is excluded, so feasibility requires that fine-tuning actually learned something.)
    \textbf{Top:} Model comparison at LoRA rank~32 (Qwen2.5-3B, Qwen3-8B, Gemma3-12B).
    \textbf{Bottom:} LoRA rank comparison on Qwen2.5-3B (ranks 8, 32, 64).
    Baseline stars show the ``$feasible@k$'' rate: the probability that at least one of $k$ independent fixed-weight runs is feasible, averaged across scenarios. Colored circles show the feasibility of a single \method run at its schedule-dependent cost.
    Mean $\pm$ std over three seeds. Higher values are better.
    }
    \label{fig:feasibility_rates}
\end{figure}

Figure~\ref{fig:feasibility_rates} reports feasibility rates, i.e., the fraction of scenarios where a method finds at least one checkpoint \emph{after step~0} that satisfies all constraints and improves target perplexity over the base model. Crucially, for this analysis, we exclude the pre-training checkpoint (step~0).

All \method schedules substantially outperform fixed-weight baselines.
In Qwen2.5-3B, the no-warmup schedule achieves the highest feasibility despite lower mean PPL reduction than warmup variants (Section~\ref{sec:results}), but not in the two larger models. Denser early updates may lead the optimizer to find higher-quality solutions when they work, but probing the loss landscape during the initial training phase may occasionally commit to weights that later violate a constraint. The no-warmup schedule avoids this early instability, offering the best feasibility at the lowest compute overhead, which may be a natural choice when reliability matters more than peak gain.

\clearpage
\section{Sensitivity to Update Frequency and Probe Length}
\label{app:alpha_sweep}

The main results use geometric schedules where the update interval doubles over training.
To assess \method's sensitivity to the frequency of weight updates, we evaluate the method with \emph{fixed} update intervals, where weights are re-estimated every $H_t$ steps throughout training. In this setting, both the interval and the probing horizon are constant and equal: $H_t \equiv H$ and $c_t \equiv H$ for all $t \in \mathcal{U}$. We sweep $H \in \{128, 256, 512\}$, corresponding to 16, 8, and 4 weight updates over the 2{,}048-step horizon.
A larger $H$ means fewer updates.

\begin{figure}[H]
    \centering
    \includegraphics[width=\textwidth]{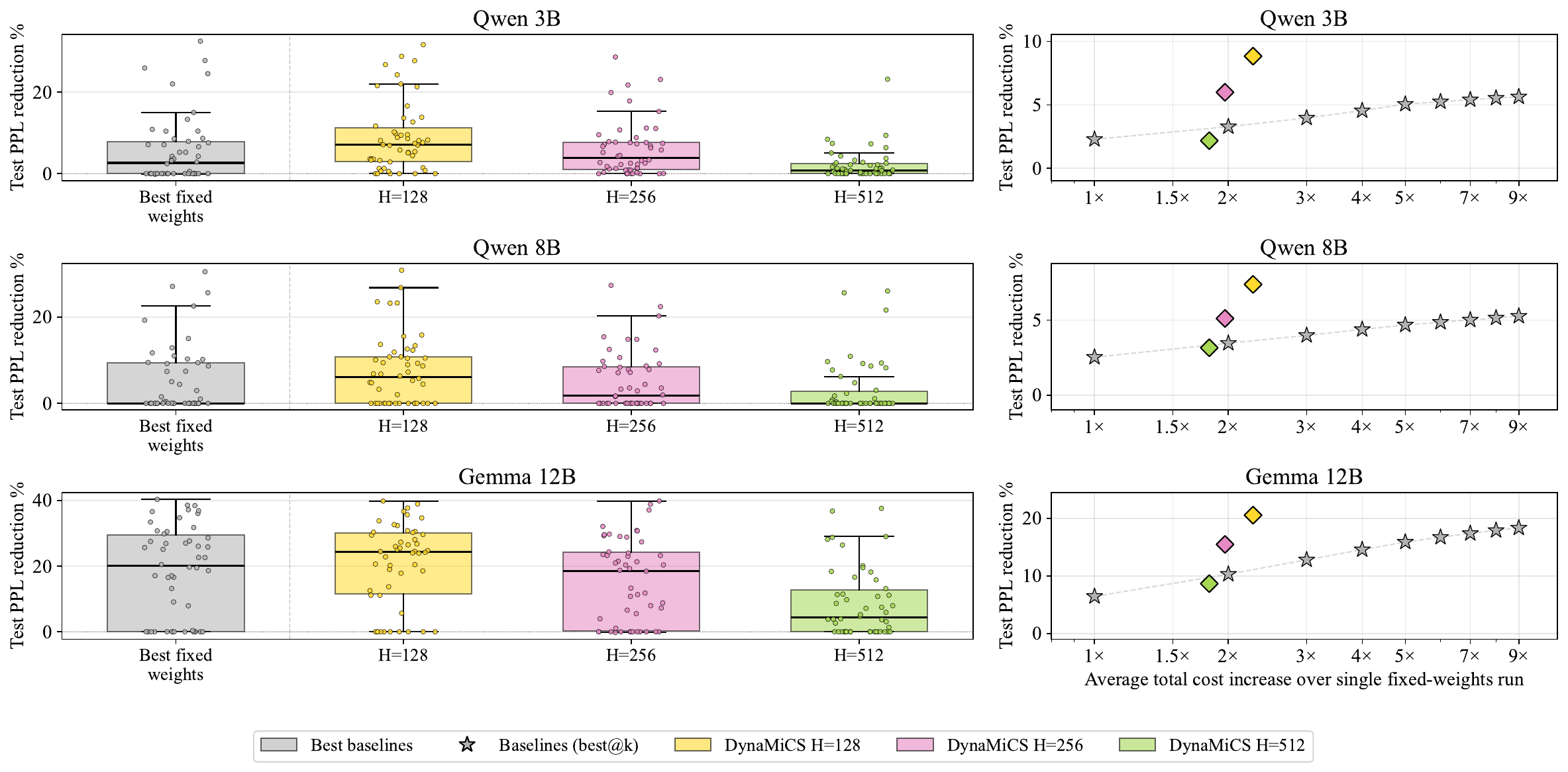}
    \caption{
    \textbf{Fixed-interval update frequency sweep: model comparison at LoRA rank~32.}
    Boxplots show per-scenario PPL reduction (one dot per scenario); for baselines, we report the oracle best across all fixed-weight configurations per scenario.
    Scatter plots show mean PPL reduction vs.\ total compute cost: baseline stars indicate the $best@k$ perplexity reduction, i.e., the best result among \textit{k} fixed-weight configurations, at the corresponding budget level, while \method diamonds show the reduction of a single adaptive run at each update interval's cost.
    Each \method configuration uses a constant reweight interval $H$ throughout training.
    }
    \label{fig:alpha_sweep_models}
\end{figure}

\begin{figure}[H]
    \centering
    \includegraphics[width=\textwidth]{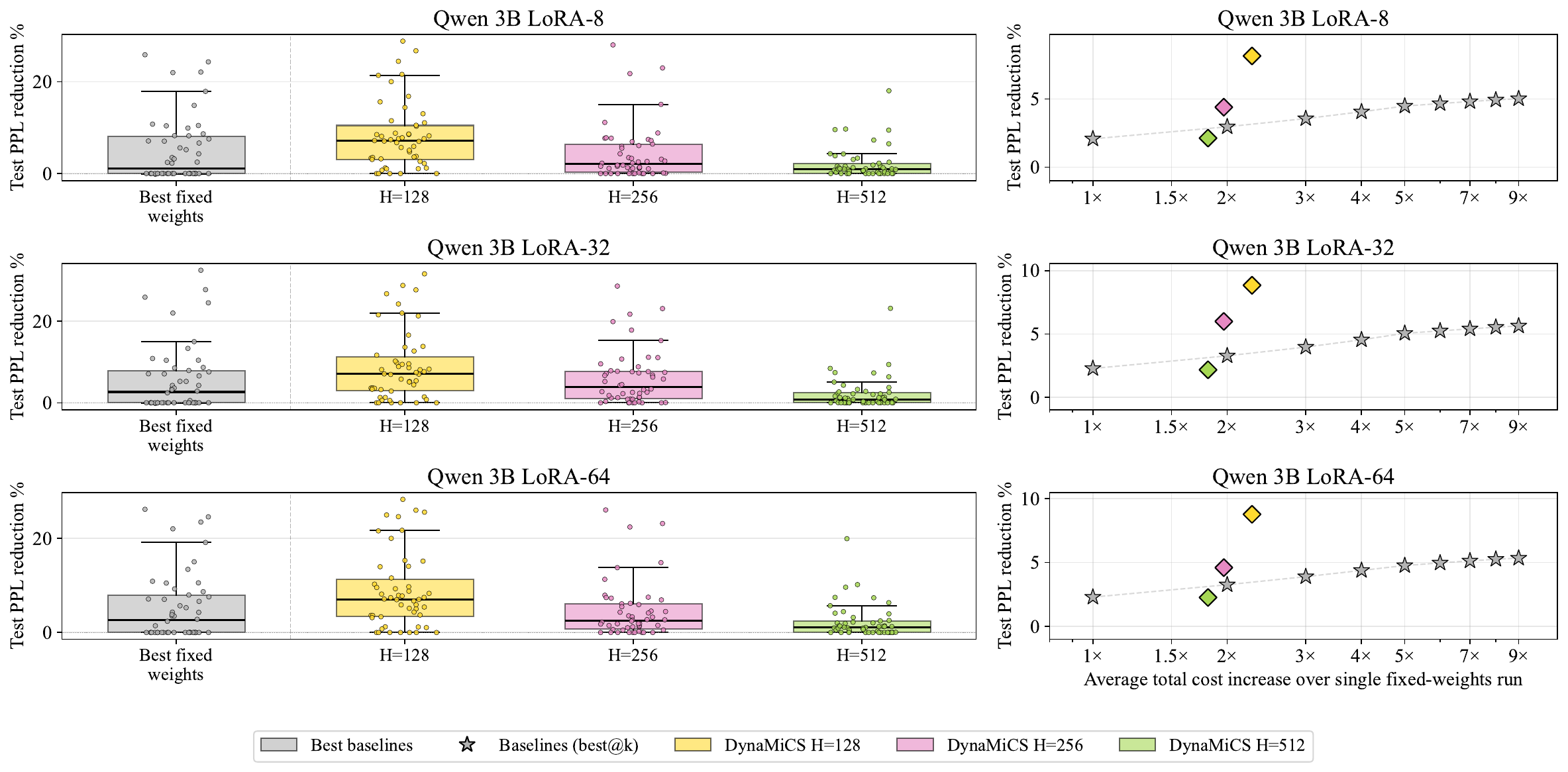}
    \caption{
    \textbf{Fixed-interval update frequency sweep: LoRA rank comparison on Qwen2.5-3B} (ranks 8, 32, 64).
    Same layout as Figure~\ref{fig:alpha_sweep_models}.
    }
    \label{fig:alpha_sweep_ranks}
\end{figure}

Figures~\ref{fig:alpha_sweep_models} and~\ref{fig:alpha_sweep_ranks} show the results.
More frequent updates ($H = 128$) yield the highest mean PPL reduction across all model configurations, consistent with the schedule results in Section~\ref{sec:results}: the optimizer benefits from responding to the evolving loss landscape more often.
Less frequent updates ($H = 512$) substantially reduce the advantage and fall below baselines.

Compared to the geometric schedules (Figure~\ref{fig:model_comparison}), fixed intervals are less cost-efficient: the geometric dense warmup achieves comparable or better PPL reduction at ${\sim}1.7\times$ cost by concentrating updates in the early training phase and spacing them out later.

\newpage

\paragraph{Cost--quality trade-off along the probe axis.}
The sweep above ties the probing budget to the update interval ($c_t \equiv H$), so it moves both at once. To separate them, we fix $H = 128$ and vary the probe length $c \in \{32, 64, 128\}$, that is $\alpha = c/H \in \{0.25, 0.5, 1\}$, on Qwen2.5-3B and Qwen3-8B (Table~\ref{tab:probe_sweep}). Very short probes are too noisy to be useful: at $c = 32$ both the perplexity reduction and the number of feasible scenarios drop on both models. Beyond $c = 64$, perplexity reduction no longer improves, while feasibility continues to rise on Qwen2.5-3B (41 to 44 of 50) and is unchanged on Qwen3-8B. Longer probes therefore trade additional compute for feasibility rather than for perplexity, and moderate probes already capture most of the useful transfer and interference signal. This ablation keeps the interval fixed throughout training rather than using a geometric schedule, so its costs are not directly comparable to the $1.4\times$--$1.7\times$ of the geometric schedules in Section~\ref{sec:results}; the $c = 128$ row coincides with the $H = 128$ configuration of the sweep above.

\begin{table}[h]
  \centering

  \begin{tabular}{cccll}
  \toprule
  Probe steps $c$ & $\alpha = c/H$ & Cost & Qwen2.5-3B & Qwen3-8B \\
  \midrule
  32  & 0.25 & $1.78\times$ & $7.9 \pm 10.4$ \; (30/50) & $6.7 \pm 10.6$ \; (21/50) \\
  64  & 0.5  & $1.94\times$ & $9.4 \pm 9.8$ \; (41/50)  & $7.7 \pm 9.1$ \; (32/50) \\
  128 & 1.0  & $2.27\times$ & $8.8 \pm 8.4$ \; (44/50)  & $7.4 \pm 7.9$ \; (32/50) \\
  \bottomrule
  \end{tabular}
    \caption{Probe-length sweep at a fixed update interval $H = 128$ (16 updates over 2{,}048 steps).
  Test perplexity reduction (mean $\pm$ std across the 50 scenarios) and the number of feasible
  scenarios, at LoRA rank 32. Costs are relative to a single fixed-weight run and are higher than those
  of the geometric schedules because the interval is held constant throughout training.}
  \label{tab:probe_sweep}
\end{table}

\clearpage
\section{LoRA Rank Sensitivity}
\label{app:rank_sensitivity}

\begin{figure}[H]
    \centering
    \includegraphics[width=\textwidth]{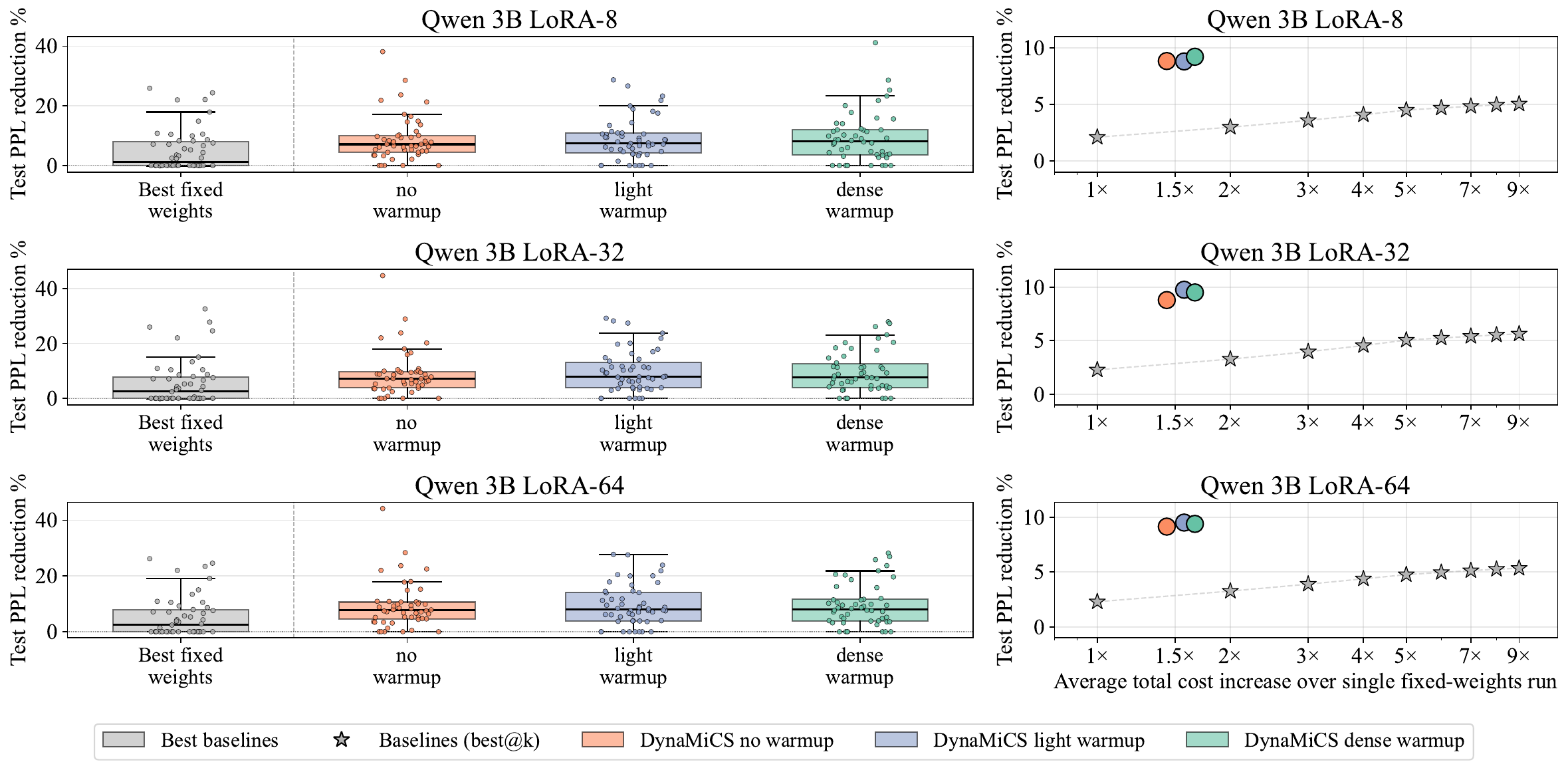}
    \caption{
    \textbf{LoRA rank comparison on Qwen2.5-3B.}
    \textbf{Left (Boxplots):} Per-scenario PPL reduction for ranks 8, 32, and 64.
    For baselines, we report the oracle best across all fixed-weight configurations per scenario.
    \textbf{Right (Scatter plots):} Mean PPL reduction vs.\ total compute cost.
    Baseline stars indicate the $best@k$ perplexity reduction, i.e., the best result among \textit{k} fixed-weight configurations, at the corresponding budget level, while \method colored circles show the reduction of a single adaptive run at its schedule-dependent cost.
    Mean $\pm$ std over three seeds.
    }
    \label{fig:rank_sensitivity}
\end{figure}

Figure~\ref{fig:rank_sensitivity} shows the effect of varying LoRA rank (8, 32, 64) on Qwen2.5-3B across all three warmup schedules.
Mean PPL reduction remains stable at ~8.7–9.8\% across ranks, and the advantage over fixed-weight baselines is consistent.

\begin{figure}[H]
    \centering
    \includegraphics[width=\textwidth]{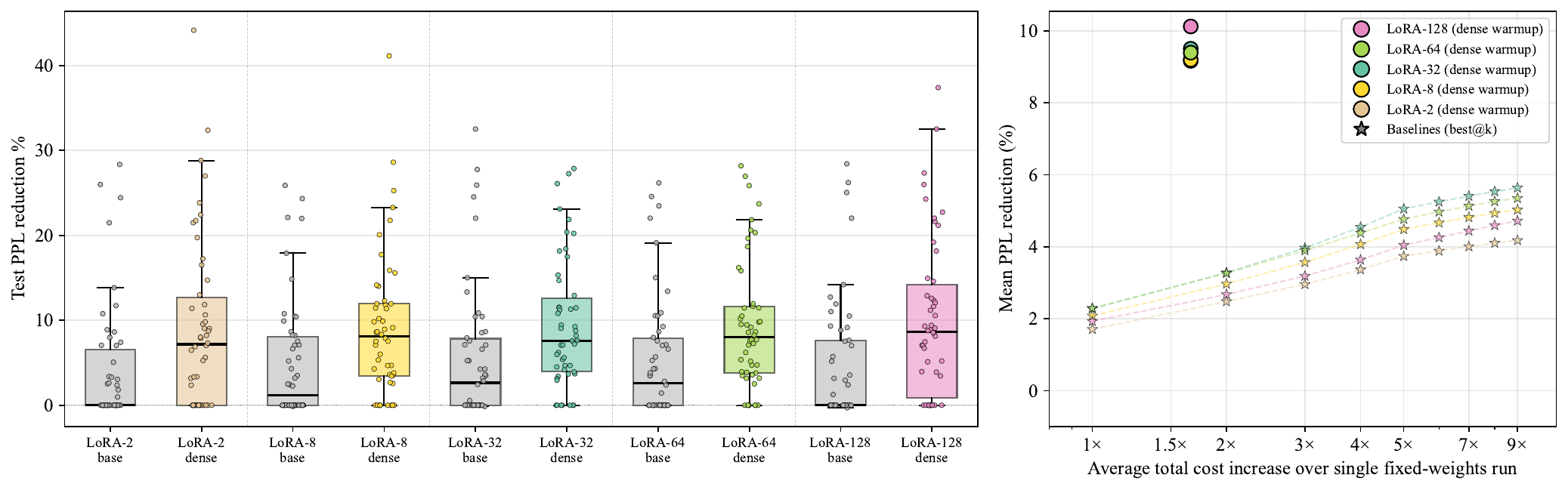}
    \caption{
    \textbf{LoRA rank comparison on Qwen2.5-3B.}
    \textbf{Left (Boxplots):} Per-scenario PPL reduction for ranks 2, 8, 32, 64 and 128.
    For baselines, we report the oracle best across all fixed-weight configurations per scenario.
    \textbf{Right (Scatter plots):} Mean PPL reduction vs.\ total compute. Baseline stars indicate the $best@k$ perplexity reduction, i.e., the best result among \textit{k} fixed-weight configurations, at the corresponding budget level, while \method colored circles show the reduction of a single adaptive run at its schedule-dependent cost.}
    \label{fig:rank_sweep_ppl}
\end{figure}

\begin{figure}[H]
    \centering
    \includegraphics[width=0.85\textwidth]{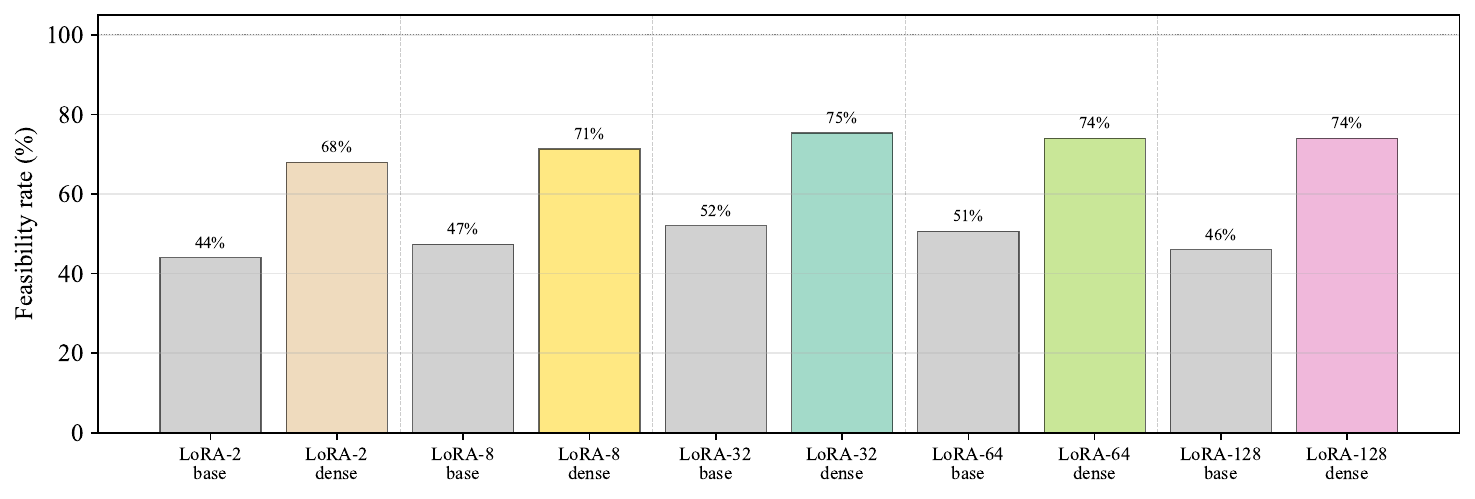}
    \caption{
    \textbf{Feasibility rate across LoRA ranks (dense warmup).}
    A run is counted as feasible only if a checkpoint after step~0 satisfies all constraints while improving target perplexity (the base model checkpoint is excluded, so feasibility requires that fine-tuning actually learned something).
    Each pair compares the fixed-weight baseline with the dense warmup schedule at a given rank. Compared to baselines,
    \method improves feasibility at every rank; performance plateaus at rank 32. Higher is better.
    }
    \label{fig:rank_sweep_feas}
\end{figure}

To verify that our default rank (32) provides appropriate adapter capacity, we additionally run a single schedule (dense warmup) across a wider range of ranks (2, 8, 32, 64, 128).
Ranks 32, 64, and 128 perform comparably in terms of feasibility and PPL reduction, indicating that additional parameters beyond rank 32 do not translate into better fine-tuning quality in our setting (Figure~\ref{fig:rank_sweep_ppl}).
Zooming into feasibility (Figure~\ref{fig:rank_sweep_feas}), ranks 2 and 8 achieve lower rates than rank 32, consistent with insufficient capacity, while ranks 32, 64, and 128 perform comparably.

\clearpage
\section{Full-parameter fine-tuning}
\label{app:full_FT}

\begin{wrapfigure}{r}{0.55\textwidth}
  \vspace{-\baselineskip}
  \centering
  \includegraphics[width=\linewidth]{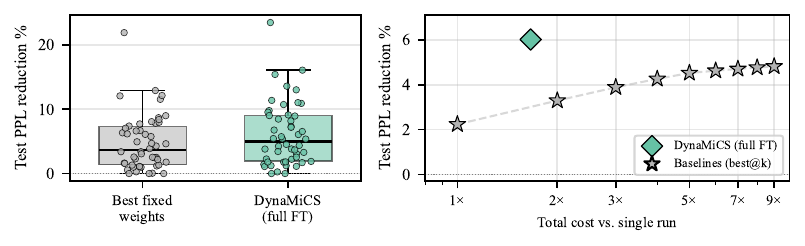}
  \caption{\textbf{Full-parameter fine-tuning (Qwen2.5-3B, all 50 scenarios).} \textbf{Left:}
  per-scenario test perplexity reduction for the best fixed-weight configuration and for a single \method run under the dense warm-up schedule; infeasible runs contribute $0\%$. \textbf{Right:} mean reduction against compute cost; stars are best@$k$ over $k$ fixed-weight runs, up to the grid size. Format as in Fig.~\ref{fig:model_comparison}.\looseness-1}
  \label{fig:fft_ppl}
  \vspace{-\baselineskip}
\end{wrapfigure}
The main experiments use LoRA, where updates are confined to a low-rank subspace. Here we ask whether \method still works when every parameter is free to move, i.e., whether the local modeling choices described in Section~\ref{sec:optimization} continue to hold in that regime.  
We rerun all 50 scenarios with full-parameter fine-tuning of Qwen2.5-3B, using the dense warm-up schedule and the same step budget, batch size and evaluation pipeline as the main experiments ($LR=10^{-6}$, decided with a sweep).

In this setup, \method reduces target perplexity by $6.0 \pm 4.8\%$, against $4.8 \pm 4.3\%$ for the best fixed-weight configuration selected per scenario, and stays above the best@$k$ curve at every budget (Figure~\ref{fig:fft_ppl}). The relative cost of \method is unchanged at ${\sim}1.7\times$, because full fine-tuning inflates the baseline run and the probing runs by the same factor. The advantage therefore persists outside LoRA.
The smaller absolute gains are explained by the smaller learning rate required for stability by full-parameter fine-tuning: the identical step budget produces less total adaptation than in its LoRA version.

\clearpage
\section{Results by Number of Constraints}
\label{app:by_constraints}

\begin{figure}[H]
    \centering
    \includegraphics[width=\textwidth]{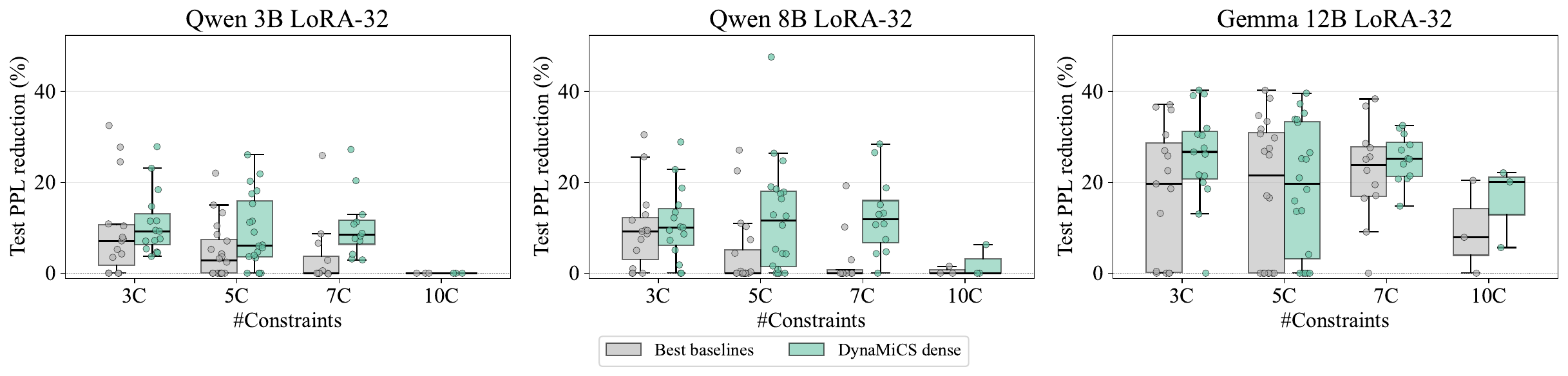}\\[0.5em]
    \includegraphics[width=\textwidth]{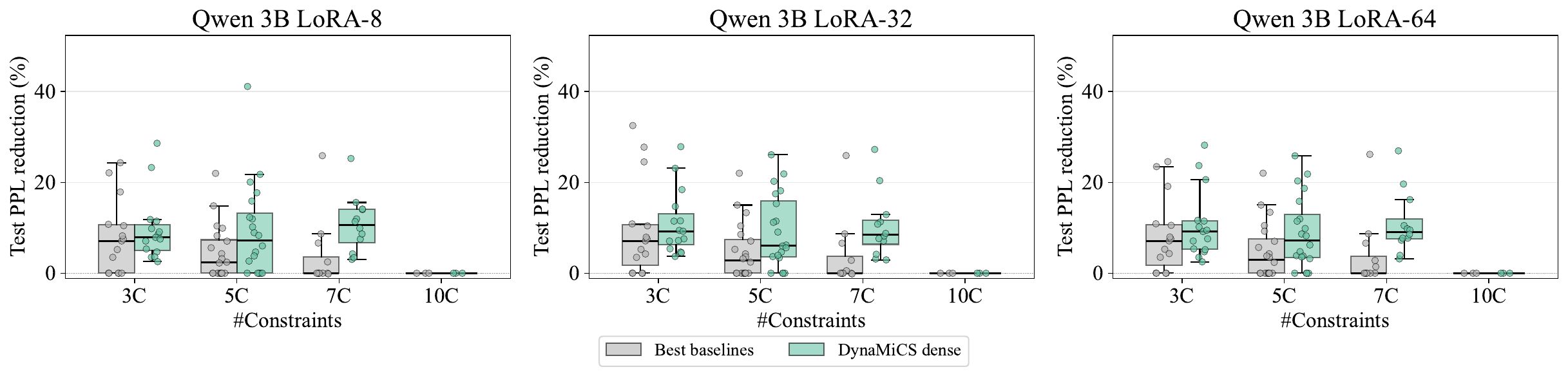}
    \caption{
    \textbf{Test perplexity reduction (\%) grouped by number of constraints} (dense warmup vs.\ oracle baselines).
    \method uses the dense warmup schedule (11 updates); baselines report the oracle best across all fixed-weight configurations.
    \textbf{Top:} Model comparison at LoRA rank~32 (Qwen2.5-3B, Qwen3-8B, Gemma3-12B).
    \textbf{Bottom:} LoRA rank comparison on Qwen2.5-3B (ranks 8, 32, 64).
    Each dot is one scenario, averaged over three seeds.
    }
    \label{fig:by_constraints_full}
\end{figure}

Figure~\ref{fig:by_constraints_full} breaks down the test perplexity reduction by the number of constraints in each scenario, using the dense warmup schedule.
\method maintains a stable advantage over baselines from 3 to 7 constraints, while baselines degrade steadily and collapse to near-zero reduction by 7 constraints.
Only at 10 constraints, the most extreme tier where the model must simultaneously preserve performance on 10 held-out domains, \method struggles as well, consistent with the inherent difficulty of satisfying that many preservation requirements at once (note, however, that solutions are still found for Qwen3-8B and Gemma3-12B).

\clearpage
\section{Constraint Violation in Infeasible Scenarios}
\label{app:violation}

Even when neither method finds a fully feasible checkpoint, \method fails more gracefully than fixed-weight baselines.
We restrict attention to the \emph{jointly infeasible} scenarios, i.e., those where both the best baseline configuration and \method (dense warmup) fail to produce any checkpoint satisfying all constraints across all seeds.
For each such scenario and method, we report the maximum constraint violation at the checkpoint closest to feasibility (i.e., the checkpoint minimizing the worst-case violation across constrained domains).

\begin{figure}[H]
    \centering
    \includegraphics[width=\textwidth]{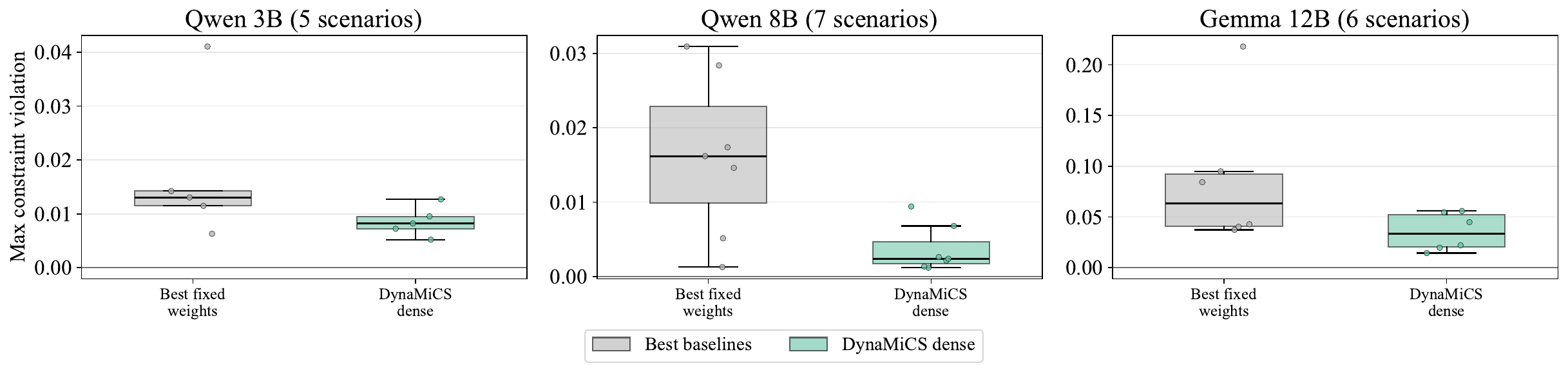}\\[0.5em]
    \includegraphics[width=\textwidth]{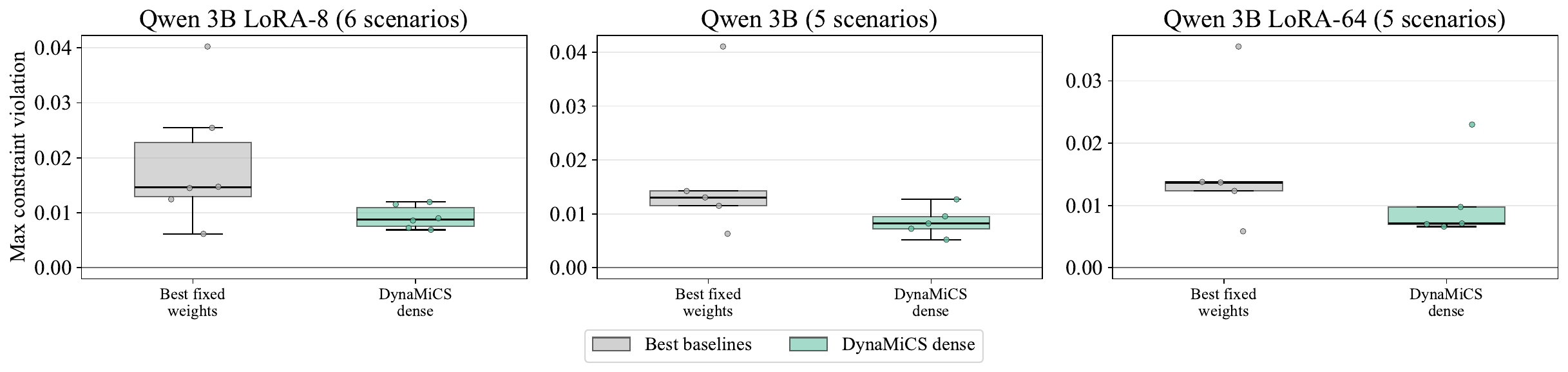}
    \caption{
    \textbf{Constraint violation in jointly infeasible scenarios} (dense warmup schedule).
    We consider the scenarios where neither \method nor any baseline configuration achieves feasibility.
    Each dot is one scenario (averaged over seeds).
    The $y$-axis shows the maximum constraint violation at the checkpoint closest to feasibility, i.e., the checkpoint minimizing the worst-case loss increase across all constrained domains.
    For baselines, we report the best (lowest violation) across all fixed-weight configurations.
    Lower is better.
    \textbf{Top:} Model comparison at LoRA rank~32 (Qwen2.5-3B, Qwen3-8B, Gemma3-12B).
    \textbf{Bottom:} LoRA rank comparison on Qwen2.5-3B (ranks 8, 32, 64).
    }
    \label{fig:violation_min}
\end{figure}

Figure~\ref{fig:violation_min} shows the results.
Across all model configurations, \method consistently achieves lower violation than the best baseline, with a more concentrated distribution closer to zero.
This indicates that the optimizer's constraint-aware weight selection reduces forgetting even in scenarios where full constraint satisfaction is unattainable, a property that baselines, which have no mechanism to respond to constraint violations, cannot match.

\clearpage

\section{Does the local linear approximation hold?}

\subsection{Fidelity of the local linear approximation}
\label{app:fidelity}

\method predicts, at each update, the loss $\hat{L}_i(w^*)$ that training on the selected mixture $w^*$ will produce $H_t$ steps later. After those $H_t$ steps we observe the realized loss $L_i(\theta_{t + H_t})$ on the same evaluation batches, which lets us measure the approximation directly rather than assume it. Per update and evaluation domain we therefore record the prediction error $e_i = L_i(\theta_{t + H_t}) - \hat{L}_i(w^*)$. The realized side is a genuine run on a mixed batch, so any cross-domain interaction that the single-domain probes could not observe is contained in this error, together with time-extrapolation error and evaluation noise. All numbers below are for Qwen2.5-3B under the dense warm-up schedule, across all 50 scenarios.

\paragraph{Error by prediction horizon.}
The probe length is capped at $c_t = \min(H_t, 128)$, so $H_t / c_t$ measures how far a prediction is used beyond the horizon on which it was fitted. Table~\ref{tab:fidelity} separates the two regimes. In seven of the nine scorable intervals the prediction is taken directly from the probe endpoint, and there the mean absolute error is 0.004 nats on losses of approximately 1.76 nats. Predicted and realized changes correlate at $r=0.93$, and a predictor that assumed no change would incur 4.3 times greater error. Error grows sharply only where the schedule extrapolates past the probe. Similarly, across 2{,}340 constrained-domain decisions, the predicted verdict on whether a constrained domain stays at or below its reference matches the realized one in 98.0\% of cases within the probed horizon, and in 83.5\% for extrapolated intervals. 

\begin{table}[h]
  \centering
  \begin{tabular}{lccc}
  \toprule
  Interval & $H_t/c_t$ & LoRA $r{=}32$ & Full FT \\
  \midrule
  $H_t \leq 128$ \; (7 of 9 intervals)          & 1      & 0.004 nats & 0.003 nats \\
  $H_t \in \{256, 512\}$ \; (2 of 9 intervals)  & 2 or 4 & 0.040 nats & 0.022 nats \\
  \bottomrule
  \end{tabular}
  \caption{Mean absolute error between predicted and realized losses, by prediction horizon.
    Qwen2.5-3B, 50 scenarios, dense warm-up schedule. Predictions are logged at the update that
    follows them, so the prediction issued at the final update ($H_t = 1024$) has no logged
    counterpart and is not scored; the same applies to the interval preceding the first update,
    which has no anchor. Nine of the schedule's eleven intervals are therefore scored.}
    \label{tab:fidelity}
\end{table}

\paragraph{Isolating the contribution of unobserved mixing.}
Interactions that the probes never see can only arise when the selected mixture spreads weight across several sources. We therefore separate updates whose weights concentrate on a single source from those that distribute across several. The error is 0.002 nats in the concentrated case and 0.007 nats in the distributed case; the 0.005-nat difference is an empirical estimate of the additional error associated with unobserved mixing interactions. This is small relative both to the loss scale of approximately 1.76 nats and to the median
distance of 0.015 nats to the constraint boundary.

\paragraph{The result is not specific to LoRA.}
Under full-parameter fine-tuning the error at $H_t/c_t = 1$ is 0.003 nats, slightly below the LoRA figure, and mixed selections are no less accurate than nearly pure ones. Full-parameter updates therefore do not degrade the local additive model in the regime where the method uses it.

\subsection{Relaxing linearity: non-linear weight updates}
\label{app:curves}

Appendix~\ref{app:fidelity} shows that the linear predictor is accurate over the horizons at which it is used. It is nonetheless worth asking whether a richer predictor would do better, particularly over the longer intervals where the linear model extrapolates past the probe. We therefore evaluate a variant that fits parametric curves to the loss trajectories observed at multiple intermediate checkpoints during slope estimation, replacing the linear extrapolation with non-linear models.
This comes at additional evaluation cost: fitting curves requires multiple checkpoints per probe (we set this number to five in our experiments), substantially increasing the overhead per weight update (see Appendix~\ref{app:linear_weight_update_costs} for a detailed cost analysis).
We use the same models as in our main experiments.

During each probing phase of $c_t$ steps on dataset $D_j$, we evaluate at five intermediate checkpoints and fit the observed loss trajectory on each evaluation dataset as a function of the normalized probe progress $u \in [0, 1]$ (with $u=0$ the pre-probe checkpoint and $u=1$ the end of the $c_t$ step probing phase), using a model selection procedure that chooses the best model among a power law with offset ($a \cdot u^b + d$) and an exponential ($a \cdot e^{-bu} + d$) using corrected Akaike Information Criterion (AICc; \citealp{Akaike, hurvich1989regression}), falling back to a linear fit when all candidates fail to converge.
Curves are fit to absolute losses (anchored at $u=0$) rather than loss deltas. A candidate mixture $\mathbf{w}$ devotes $w_j H_t$ of the interval's steps to $D_j$, i.e.\
$u_j = w_j H_t / c_t$ in normalized units, and the fitted curves combine additively:
$\hat{L}_i(\mathbf{w}) = L_i(\theta_t) + \sum_{j} \big[ f_{ij}(u_j) - f_{ij}(0) \big]$. The optimizer solves the constrained problem~\eqref{eq:main_problem} using these predictions at the target horizon.
We evaluate this variant with the same fixed-interval schedules as Appendix~\ref{app:alpha_sweep}, sweeping $H \in \{128, 256, 512\}$ with probing budget $c = H$. 

\begin{figure}[H]
    \centering
    \includegraphics[width=\textwidth]{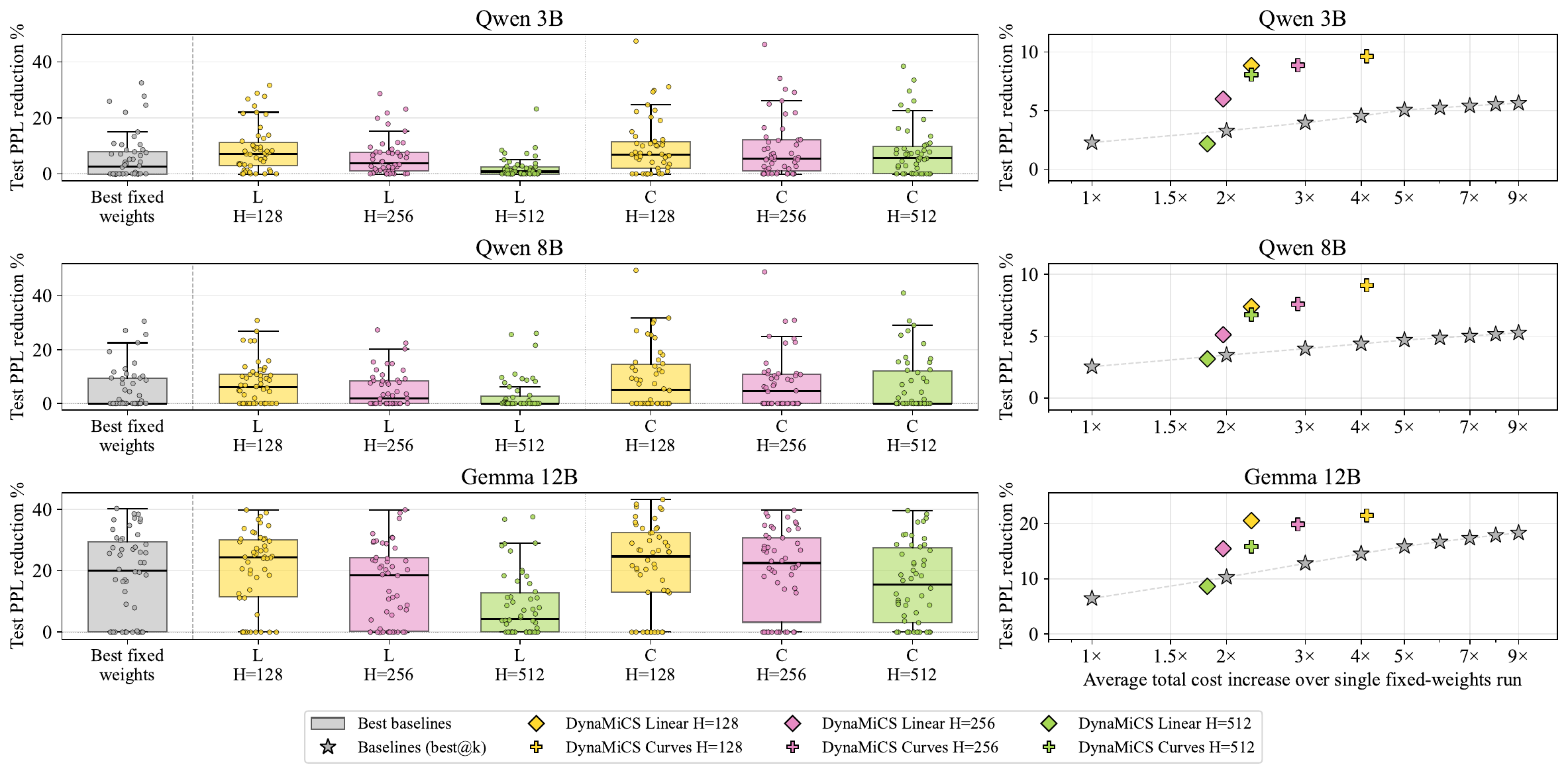}
    \caption{
    \textbf{Curve-fitting predictor: model comparison at LoRA rank~32.}
    Same layout as Figure~\ref{fig:alpha_sweep_models}.
    Each \method configuration uses a constant reweight interval $H$ with probing budget $c = H$ and a non-linear predictor (fitted on five intermediate evaluations steps).
    }
    \label{fig:alpha_curves_models}
\end{figure}

\newpage
\begin{figure}[H]
    \centering
    \includegraphics[width=\textwidth]{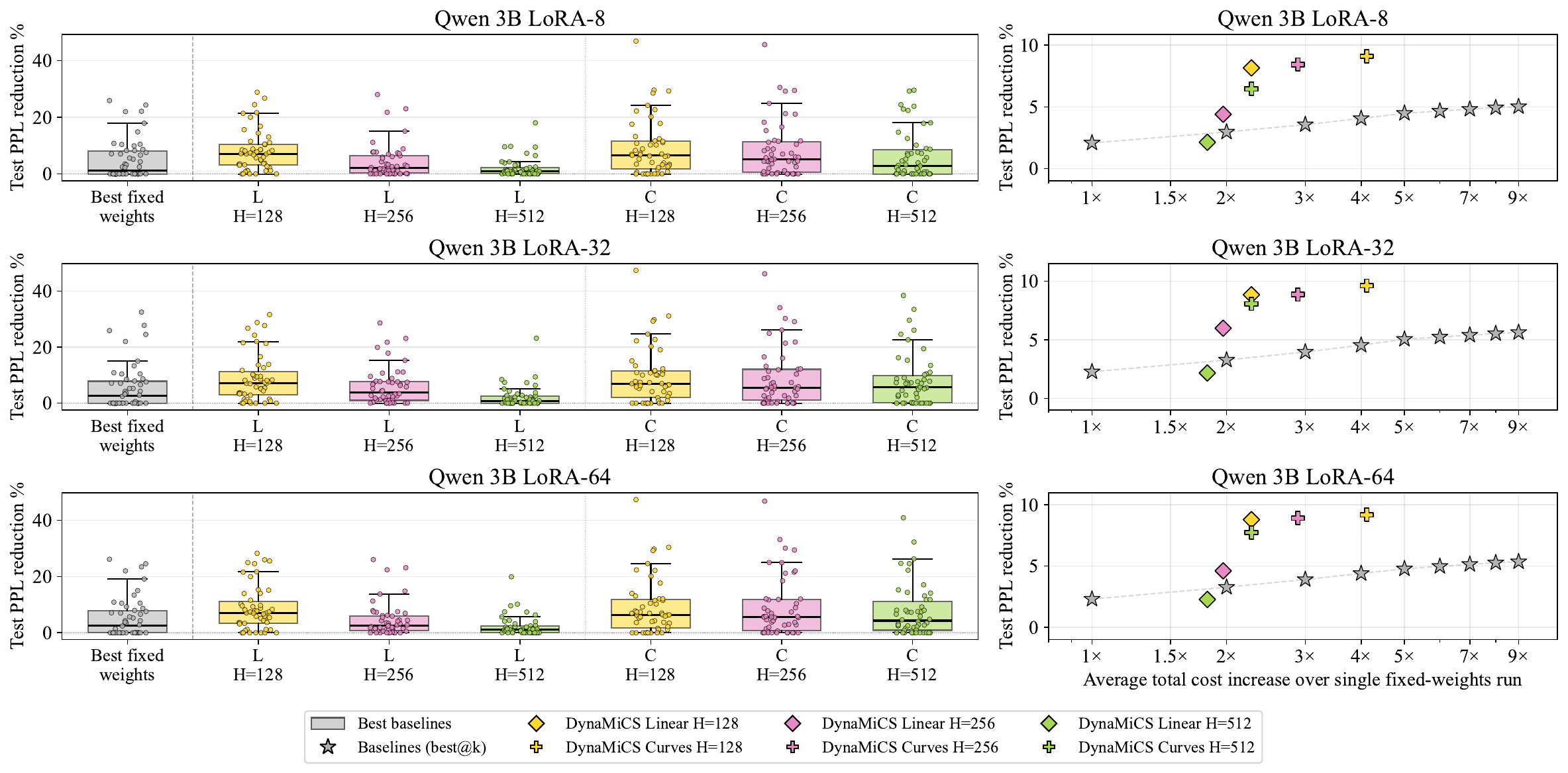}
    \caption{
    \textbf{Curve-fitting predictor: LoRA rank comparison on Qwen2.5-3B} (ranks 8, 32, 64).
    Same layout as Figure~\ref{fig:alpha_sweep_models}.
    }
    \label{fig:alpha_curves_ranks}
\end{figure}

Figures~\ref{fig:alpha_curves_models} and~\ref{fig:alpha_curves_ranks} show the results.
The curve-fitting variant is less sensitive to large update intervals than its linear counterpart: at $H = 512$, the linear predictor degrades sharply, whereas curves retain a larger fraction of the reduction achieved at $H = 128$.
This is consistent with the non-linear predictor's ability to describe longer horizons more accurately.
However, this robustness comes at a higher compute cost per update (shifted rightward on the cost axis), and does not translate into an overall advantage.
These results suggest that the additional compute is better spent on tighter update schedules, such as the geometric schedules in Section~\ref{sec:schedules}, that concentrate updates early in training, when the loss landscape changes most rapidly, rather than on richer per-probe predictors.

While further exploration of non-linear models remains a promising direction for longer probing horizons or strongly non-linear regimes, for short probing budgets the linear approximation offers the best cost-performance trade-off.
This is consistent with Appendix~\ref{app:fidelity}: since the linear predictor's error is already small at the horizons where the schedule places most of its updates, and well inside the margin separating a feasible mixture from an infeasible one, additional predictive accuracy there has little left to correct.

\clearpage
\section{Additional Baselines}
\label{app:baselines}

\subsection{Gradient Alignment}
\label{app:gradient_alignment}

\paragraph{Formulation.}
Gradient alignment~\citep{ren2019learningreweightexamplesrobust,yu2020gradientsurgerymultitasklearning,fan2024dynamicgradientalignmentonline} estimates the slope matrix without performing any fine-tuning steps.
Instead of probing each dataset for $c_t$ steps and observing loss changes, it computes average gradients over $n_b$ batches and uses their inner products as slope estimates:
\begin{equation}
\label{eq:grad_align_slope}
S_{ij}^{\text{GA}} = -\eta\,\langle g_i,\, d_j \rangle,
\end{equation}
where $g_i$ is the average gradient of evaluation dataset $E_i$, $d_j$ is the Adam update direction for fine-tuning dataset $D_j$ (bias-corrected first moment divided by the square root of the second moment), and $\eta$ is the learning rate.
The intuition is that if the Adam step for dataset $D_j$ aligns with the direction that decreases $L_{i}$, training on $D_j$ should reduce the loss on $E_i$.

The matrix is re-estimated at every update step $t \in \mathcal{U}$ and passed to the same constrained optimization (Equation~\ref{eq:penalized}) used to derive \method's mixture weights, so this baseline is dynamic in the same sense as \method and differs from it only in how the slope matrix is obtained.
Its cost is driven by the number of batches used to average gradients per dataset (in our experiments, we use 200); since no multi-step probing is required, this is comparable to \method configurations with tight update schedules or larger evaluation budgets: the difference in compute is not the deciding factor.

\paragraph{Approximation issue.}

The key limitation of gradient alignment is that Equation~\ref{eq:grad_align_slope} provides an \emph{instantaneous} rate of change that is difficult to translate into loss predictions with proper scale over a horizon $H_t$.
One interpretation of $S_{ij}^{\text{GA}}$ is that it is a Taylor approximation of the loss drop on $E_i$ after one step of Adam on $D_j$:
$$
S_{ij}^{\text{GA}} \simeq L_{i}(\theta'_t) - L_{i}(\theta_t), \text{ with } \theta'_t = \theta_t - \eta d_j.
$$
Using it to predict the loss after $H_t$ steps leads to accumulating errors due to i) the Taylor expansion higher order terms, ii) the stochastic nature of the algorithm not taken into account. 

In contrast, \method's multi-step probing observes actual loss trajectories and directly measures the quantity of interest ---the loss change after $c_t$ steps--- making the linear extrapolation to larger horizon better grounded.

\begin{figure}[H]
    \centering
    \includegraphics[width=\linewidth]{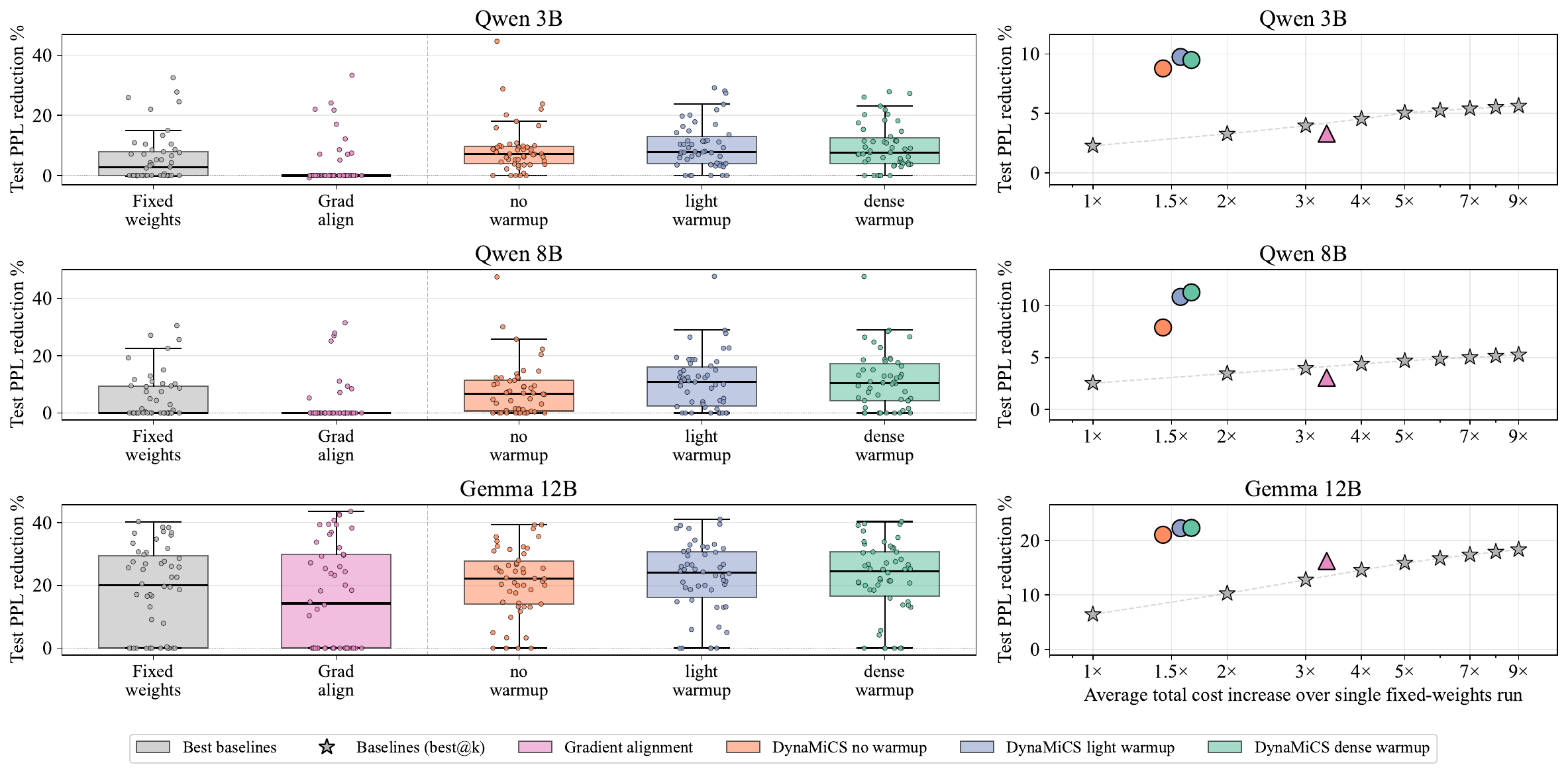}
    \caption{
    \textbf{Gradient alignment vs.\ \method across three models.} (LoRA rank 32, 50 scenarios)
    Boxplots show per-scenario test PPL reduction (one dot per scenario); for fixed-weight baselines, we report the oracle best across all configurations per scenario.
    Scatter plots show mean PPL reduction vs.\ total compute cost: Baseline stars indicate the $best@k$ perplexity reduction, i.e., the best result among \textit{k} fixed-weight configurations, at the corresponding budget level, while \method circles and gradient alignment triangles show the reduction of a single adaptive run at its cost.\looseness-1
    }
    \label{fig:grad_align_results}
\end{figure}

\paragraph{Results.}
Figure~\ref{fig:grad_align_results} compares gradient alignment against \method and fixed-weight baselines across three models.
Gradient alignment consistently underperforms \method.
The poor performance is consistent with the scale issue described above: while gradient alignment can identify beneficial \emph{directions} (which datasets help which), it struggles to produce slope magnitudes that are calibrated for the constraint optimization and horizon, leading to either overly conservative or overly aggressive weight allocations.

\subsection{Embedding Similarity}
\label{app:embeddings}

\paragraph{Formulation.}

Like gradient alignment, this baseline keeps \method's constrained optimizer and changes only where the slope matrix comes from, testing a natural intuition: that fine-tuning datasets resembling a target should be upweighted. With this goal, we embed every domain with the base model, mean-pooling the last hidden state over $n_b$ batches and centering globally, and take the cosine similarity $\mathrm{sim}_{ij}$ between evaluation domain $E_i$'s validation data and fine-tuning dataset $D_j$'s training data.
 
We establish a threshold to turn this one-signed quantity into a signed influence:
  \begin{equation}
  \label{eq:sim_slope}
  S_{ij}^{\text{sim}} = -(\mathrm{sim}_{ij} - \tau)\, s ,
  \end{equation}
so that similarity above $\tau$ is read as potentially beneficial transfer and similarity below it as possible interference, with $s$ a scale factor. No training or probing takes place, only forward passes to embed the domains at step 0. The matrix is handed to the same weight updater and constrained solve as \method, only once at step~0, so the resulting mixture is held fixed for the rest of training. We sweep $\tau \in \{0.1, 0.3, 0.5\}$ with $s = 1$ and $n_b = 8$.

\paragraph{Approximation issue.}
The key limitation of this baseline is that a similarity cannot be converted into a loss change in any principled way. Equation~\ref{eq:sim_slope} must map a dimensionless quantity onto the nats in which Equation~\ref{eq:penalized} compares predicted losses against reference levels, and nothing in the signal fixes $\tau$ or $s$: similarity, instead, records how related two domains are, not how much training on one moves the loss on the other. It is also blind to how much data each domain contains and, being computed once from the base model, cannot track interactions that evolve.

\paragraph{Results.}
Across the three thresholds the mixtures land in a narrow band: mean test perplexity reduction of $2.02\%$ ($\tau{=}0.1$), $2.29\%$ ($\tau{=}0.3$) and $1.48\%$ ($\tau{=}0.5$), feasible in 15, 19 and 17 of the 50 scenarios respectively. Even the best threshold therefore recovers less than a quarter of \method's $9.50\%$ and satisfies the constraints in fewer scenarios than the fixed-weight grid (Section~\ref{app:baselines_summary}). These results are consistent with the embedding similarity lacking the signal the optimizer needs.

\subsection{DoReMi}
\label{app:doremi}

\paragraph{Formulation.}
DoReMi~\citep{doremi} learns domain weights with a small proxy model before the main run, returning a single mixture, which is then held fixed throughout training (see Appendix \ref{app:related-work} for details). We maintain DoReMi's two-phase structure and its published hyper-parameters, adapting it from the pre-training setting it was
designed for to fine-tuning. In Phase~A a reference model (Qwen2.5-0.5B with LoRA $r{=}32$) is fine-tuned for 2{,}048 steps on the uniform mixture. In Phase~B a proxy model of the same size
trains for 2{,}048 steps while its domain weights are updated every 32 steps by exponentiated gradient on the clipped excess loss of proxy against reference ($\eta{=}1$, smoothing $c{=}10^{-3}$, excess losses measured on 16 held-out batches per domain), a Group-DRO objective on the worst-case excess loss across the fine-tuning domains. The mixture we transfer to Qwen2.5-3B is the time-average of those weights along the proxy trajectory.

\paragraph{Objective mismatch.}
DoReMi optimizes a different objective from ours: it minimizes the worst-case excess loss across the datasets it trains on, while we improve some evaluation domains subject to keeping others from degrading. Nothing in that objective distinguishes those two roles, so no reweighting of it can express the requirement that a particular loss stay below a reference level: feasibility is not a quantity DoReMi optimizes. The method also returns a single static mixture, so it cannot react to transfer and interference as they evolve.

\paragraph{Results.}
The transferred mixture reaches a mean test perplexity reduction of $2.6 \pm 6.3\%$ and is feasible in 10 of the 50 scenarios, against $9.50\%$ for \method (Section~\ref{app:baselines_summary}). It also falls below the fixed-weight grid, which is expected: the grid result is the best of \textit{k} mixtures, picked per scenario as an oracle, based on the results each fixed-weight recipe achieved. In contrast, DoReMi commits to a single mixture derived from the fine-tuning datasets alone, without reference to which evaluation domains the scenario is trying to improve.

\subsection{Skill-It}
\label{app:skillit}

\paragraph{Formulation.}
Skill-It~\citep{skillit} is the structural precursor of our slope matrix: it learns a non-negative skill-influence graph once,
then reweights the mixture by multiplicative weights. It therefore replaces both parts of \method: the estimator of cross-domain effects and the update rule that consumes it. To instantiate it, we
build the graph with Skill-It's Algorithm~3, the variant its authors prescribe when training and evaluation skills do not coincide. Algorithm~3 measures exactly what our probe measures, the loss
change on each evaluation domain after training on a single fine-tuning dataset in isolation, so we reuse our own probing machinery and read the graph off the slope matrix,
\begin{equation}
A \;=\; \max\!\big(0,\, -c_0\,\mathbf{S}(0)\big) \;\in\; \mathbb{R}^{M \times N},
\end{equation}
where $c_0 = 150$ is the probing budget, so that $c_0\,\mathbf{S}(0)$ is the total loss change over the probe rather than the per-step rate of~\eqref{eq:slope}, and $A_{ij} > 0$ measures how much training on $D_j$ helps evaluation domain $E_i$. The clamp is Skill-It's own: its graph learner records an edge only where the probe decreases
the loss (Algorithm~3, line~6), so measured interference never enters $A$. Following the paper, we estimate the graph once, with $150$ probing steps, and freeze it for the rest of training. Weights are initialized as in Skill-It's Algorithm~1 (line~3), $w_j^{(1)} \propto \exp\!\big(\eta \textstyle\sum_{i} A_{ij}\big)$, and are updated at each $t \in \mathcal{U}$ by
\begin{equation}
w_j^{(t+1)} \;\propto\; w_j^{(t)}\,
\exp\!\Big(\eta \textstyle\sum_{i} A_{ij}\,\bar{L}_i\Big),
\end{equation}
where $\bar{L}_i$ is the running mean of the losses observed on $E_i$ and $\eta$ is Skill-It's update rate. Training budget, update schedule $\mathcal{U}$ and evaluation cadence are those of \method to allow fair comparison.

Skill-It has no natural way to express a constraint, that is, a domain whose performance must not degrade.  It is nonetheless interesting to ask whether this distinction can be incorporated into  the method. To this aim, we add a weight $\lambda$ on the constrained-domain rows of $A$: at $\lambda = 0$ constraints are ignored and only targets drive the update; at $\lambda = 1$ they are treated as additional targets, which is Skill-It as published; $\lambda > 1$ emphasizes them further. We sweep $\lambda \in \{0, 0.1, 1, 10\}$ against Skill-It's update rate $\eta \in \{0.05, 0.1, 0.2, 0.5, 1, 10\}$ and select the best of the resulting configurations independently per scenario, an oracle choice that favors the baseline.

\paragraph{Objective mismatch.}
Skill-It's update rule treats every evaluation domain as a loss to be minimized, and no reference level informs the objective: what drives the update is the magnitude of each loss, not its position relative to $L_i^{\text{ref}}$. The $\lambda$ knob does not supply one, since it rescales the
constrained rows of $A$ without introducing any level at which a loss counts as acceptable, and the rows it rescales encode positive transfer only. If a constrained domain is actively damaged by a fine-tuning dataset $D_j$, Skill-It's optimization can therefore react only by up-weighting the
datasets that were measured to help that domain. Its cell for $D_j$ is zero, so the damaged domain neither raises nor lowers $D_j$'s weight, and $D_j$ continues to be promoted by the target domains it does help. Raising $\lambda$ dilutes $D_j$ only indirectly, by promoting the constraint's helpers faster: it never acts on the source of the damage directly.

\paragraph{Results.}
Across the $\lambda \times \eta$ grid, the per-scenario best configuration reaches a mean test perplexity reduction of $3.5 \pm 7.8\%$ and is feasible in 16 of the 50 scenarios, against $9.50\%$ and 40 of 50 for \method (Section~\ref{app:baselines_summary}). This is the best of 24 configurations selected per scenario after the fact, whereas \method is reported at a single setting held fixed across all scenarios. As with the other baselines, this is not a shortcoming of Skill-It on the problem it was designed
for: it selects a mixture that accelerates learning across a set of skills, and a graph that cannot represent forgetting offers nothing to optimize against when a capability must be
preserved.

\subsection{Aioli}
\label{app:aioli}

\paragraph{Formulation.}
Aioli~\citep{aioli} is the closest prior method to \method. We replace both of its components: the estimator of cross-domain effects and the update rule that consumes it. Its estimator infers the parameters of a linear mixing law from the loss mixture trajectory of the training run itself, interleaving short chunks on near-one-hot mixtures ($1-\epsilon$ on one domain, $\epsilon$ spread over the rest, with $\epsilon = 0.1$); its updater then takes one exponentiated-gradient step on the simplex. We inherit both, keeping Aioli's structure and defaults, and match its measurement budget to ours exactly. Following the paper, updates land at equal intervals rather than on our geometric schedule. 
We run two versions of this baseline: one loyal to Aioli's published objective, which averages the loss over all evaluation domains, and a target-aimed version that directs the update at the target domains. We sweep the exponentiated-gradient step size, the one hyper-parameter its own sensitivity analysis
identifies as influential. Our implementation deviates from the paper in one respect: we restore the model after each measurement stretch, whereas Aioli never rewinds.

\paragraph{Objective mismatch.}
Aioli minimizes an aggregate loss over the evaluation domains, without constraints. It therefore has no way to express that one particular loss must stay below a reference level, and no reweighting of its objective supplies one: both variants we run are legitimate choices of the group-specific $b^t$ vector of its Lemma~1, but neither can encode a reference level, so aiming the update at the targets pushes harder on them but leaves the remaining domains unprotected, while averaging over all of them dilutes the targets. Its estimator, by contrast, transfers to our setting unchanged: the linear solve is over the measurement mixtures, so each evaluation domain's row is independent and a rectangular matrix over two decoupled index sets---including targets with no training data of their own---costs nothing to support.

\paragraph{Results.}
Aioli reaches a mean test perplexity reduction of $3.1 \pm 7.8\%$ and is feasible in 9 of the 50 scenarios, against $9.50\%$ for \method (Section~\ref{app:baselines_summary}). Neither raising the step size nor switching between the aggregate and the target-aimed objective moves that feasibility count, and the two step sizes produce visibly different weight trajectories while landing within $0.02$ percentage points of each other, so the shortfall is structural rather than a matter of tuning. This is not a shortcoming of Aioli on the problem it was designed for: it optimizes an unconstrained aggregate loss, and a method with no representation of knowledge preservation constraints has no machinery to satisfy one.

\vspace*{-0.15cm}
\subsection{Baselines Summary}
\label{app:baselines_summary}

\begin{table}[t]
\centering
\vspace*{-0.29cm}
\begin{tabular}{lcc}
\toprule
Method & Mean test PPL reduction & Feasible \\
\midrule
Best fixed weights       & $6.9 \pm 9.2\%$ & 29/50 \\
Gradient alignment              & $3.3 \pm 7.5\%$ & 12/50 \\ 
Embedding similarity     & $2.3 \pm 5.0\%$ & 19/50 \\
DoReMi                          & $2.6 \pm 6.3\%$ & 10/50 \\
Skill-It               & $3.5 \pm 7.8\%$ & 16/50 \\
Aioli                   & $3.1 \pm 7.8\%$ & 9/50 \\
\midrule
\method                         & $9.5 \pm 7.5\%$ & 40/50 \\
\bottomrule
\end{tabular}

\caption{\textbf{Baselines on the 50 scenarios} (Qwen2.5-3B, LoRA $r{=}32$). Feasibility follows Section~\ref{sec:metrics}: infeasible runs contribute $0\%$ to the perplexity reduction. Spreads are population standard deviations across the 50 scenarios. Gradient alignment, DoReMi and \method each have a single configuration; for the fixed-weight grid and Skill-It we report the best configuration
selected per scenario, and for embedding similarity and Aioli the best single configuration applied to all 50 scenarios. The fixed-weight and \method rows are averaged over three seeds; additional baselines use a single seed.\looseness-1}
\label{tab:baselines_summary}
\end{table}

Table~\ref{tab:baselines_summary} collects the baselines on the setting they all share, Qwen2.5-3B at LoRA $r{=}32$ across the 50 scenarios. They fall into two groups. Gradient alignment and embedding similarity keep \method's constrained optimizer and replace only the estimator of cross-domain effects, so they isolate how much the probed slope matrix contributes; DoReMi, Skill-It and Aioli replace the objective as well, and none of the three can express a preservation constraint in the first place. The fixed-weight and \method rows are those reported in the main text and are averaged over three seeds. The additional baselines reported in this appendix use a single seed.

\clearpage
\section{Choosing the Adaptation--Retention Trade-off}
\label{app:unconstrained}

The reference levels $L_i^{\text{ref}}$ are set by the experimenter, so how much \textit{learning} to give up in order to protect a capability is a tradeoff that the experimenter can control. To quantify this, we run the two extremes of that choice on all 50 scenarios (Qwen2.5-3B, LoRA $r{=}32$, dense warm-up schedule): strict preservation (each $L_i^{\text{ref}}$ set to its step-0 value, i.e. zero allowed degradation), and no constraints at all. In practice, in the unconstrained setup the target/constraint partition is removed and the objective is the target loss alone, while the constrained domains are still evaluated only in order to measure their degradation.

\begin{table}[h]
  \centering
  
  \begin{tabular}{lcc}
  \toprule
   & Mean target-PPL reduction & Feasible scenarios \\
  \midrule
  Unconstrained (target loss only) & $16.7 \pm 9.0\%$ & 11/50 \\
  Constrained (zero tolerance)     & $9.5 \pm 7.5\%$  & 40/50 \\
  \bottomrule
  \end{tabular}
  \caption{The two extremes of the trade-off (Qwen2.5-3B, LoRA $r{=}32$, 50 scenarios). Feasibility follows Section~\ref{sec:metrics} for both setups, though for the unconstrained arm it is a post-hoc audit, since
  nothing in that run optimizes for it. The perplexity reduction for the constrained setup is read at its best feasible checkpoint and contributes $0\%$ where none exists, while for
  the unconstrained setup is read at the best checkpoint its own objective selects.}
  \label{tab:trade_off}
\end{table}

Strict preservation costs about 7 percentage points of target-perplexity reduction. However, without the constraints, 44 of the 50 scenarios regress a protected capability. Which of the two is preferable depends on the application, and the reference levels let the experimenter place the run anywhere between them.

Overall, data mixing cannot dissolve the underlying conflict between adaptation and retention: if a dataset carries both transfer signal and interference, no allocation of mixture weights can solve the problem, since the conflict lives in the gradients. \method does not remove this conflict; it makes it explicit, measurable, and steerable by the practitioner.

\clearpage
\section{Expressing Targets and Constraints in Terms of Accuracy}
\label{app:accuracy_experiments}

\subsection{Accuracy as a target, per benchmark}

\begin{figure}[H]
  \centering
  \includegraphics[width=\linewidth]{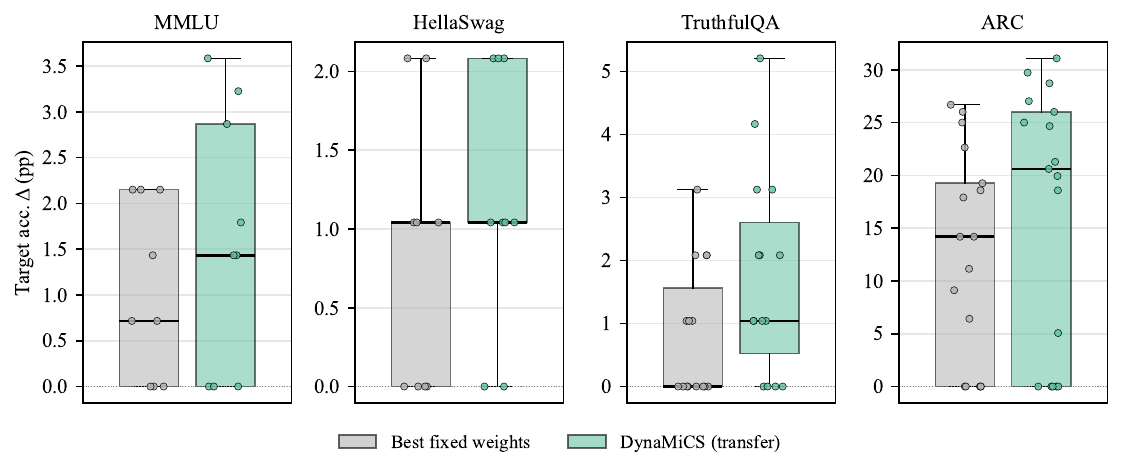}
  \caption{\textbf{Held-out accuracy target, by target benchmark (Qwen2.5-3B, $r{=}32$).} Per-scenario change in target accuracy (pp) for the best fixed-weight configuration and for a single \method run, split by which benchmark was promoted to target. Each panel contains the scenarios in which that
  benchmark is the target and has its own vertical scale. Higher is better.}
  \label{fig:accuracy_by_benchmark}
\end{figure}

Section~\ref{sec:results} reports the held-out accuracy target aggregated over the 50 scenarios. Because the four benchmarks differ in how much of their performance is reachable through transfer from the available fine-tuning data, Figure~\ref{fig:accuracy_by_benchmark} breaks the same runs down by which benchmark was promoted to target. On average, \method improves over the best fixed-weight configuration on all four benchmarks.

\subsection{Accuracy as a constraint}

\begin{wrapfigure}{r}{0.55\textwidth}
  \vspace{-\baselineskip}
  \centering
  \includegraphics[width=\linewidth]{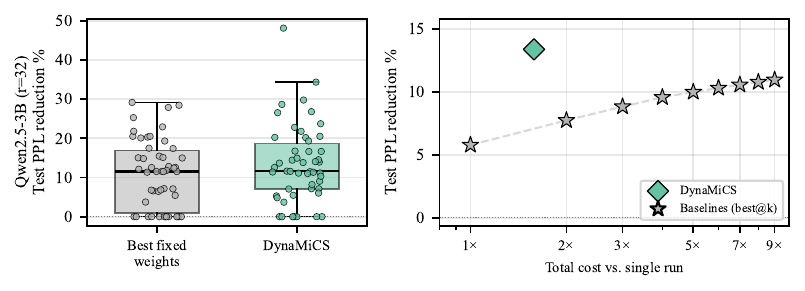}
  \caption{\textbf{Accuracy as a constraint (Qwen2.5-3B, $r{=}32$).} \textbf{Left:} per-scenario
  test-perplexity reduction on the target domains, for the best fixed-weight configuration and for a
  single \method run. \textbf{Right:} mean reduction against compute cost; stars are best@$k$ over $k$
  fixed-weight runs. Format as in Fig.~\ref{fig:model_comparison}.}
  \label{fig:accuracy_constraints}
  \vspace{-\baselineskip}
\end{wrapfigure}
Accuracy can also define a capability to be preserved rather than improved. We rerun all 50 scenarios with the benchmark constraints expressed as accuracy instead of loss, measured as the length-normalized log-likelihood of the choice text: a benchmark is preserved when its
accuracy does not fall below its pre-fine-tuning value. Constraints on non-benchmark domains remain loss-based, so scenarios mix the two metrics and \method satisfies them jointly. Everything else follows the main experiments. On Qwen2.5-3B, \method reduces target perplexity by $12.9 \pm 10.1\%$ against $10.0 \pm 8.8\%$ for the best fixed-weight configuration.

\clearpage
\section{Evolution of weight allocation}
\label{app:weight}

\begin{figure}[H]
    \centering
    \includegraphics[width=\textwidth]{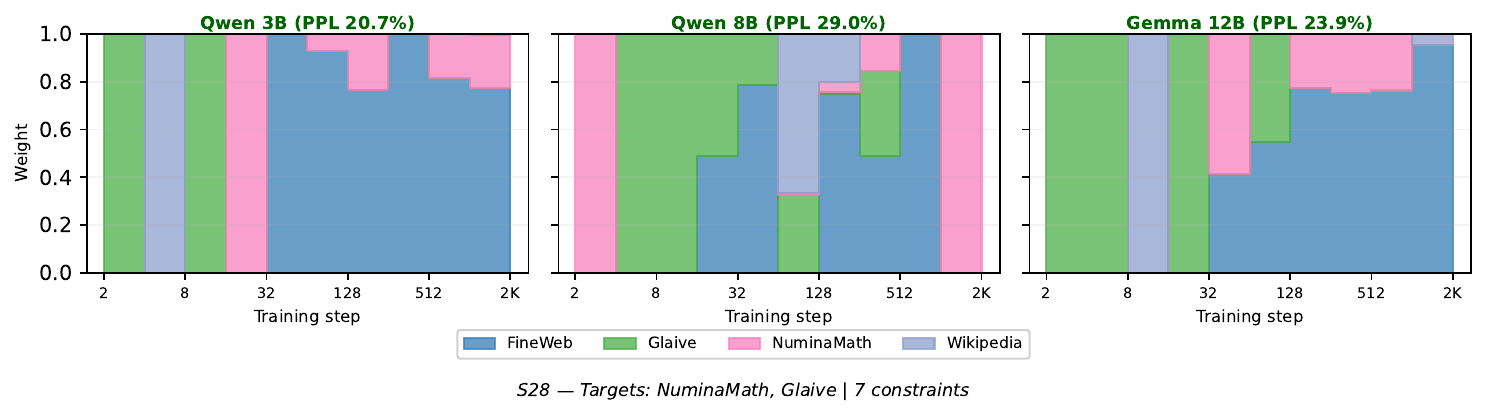}
    \includegraphics[width=\textwidth]{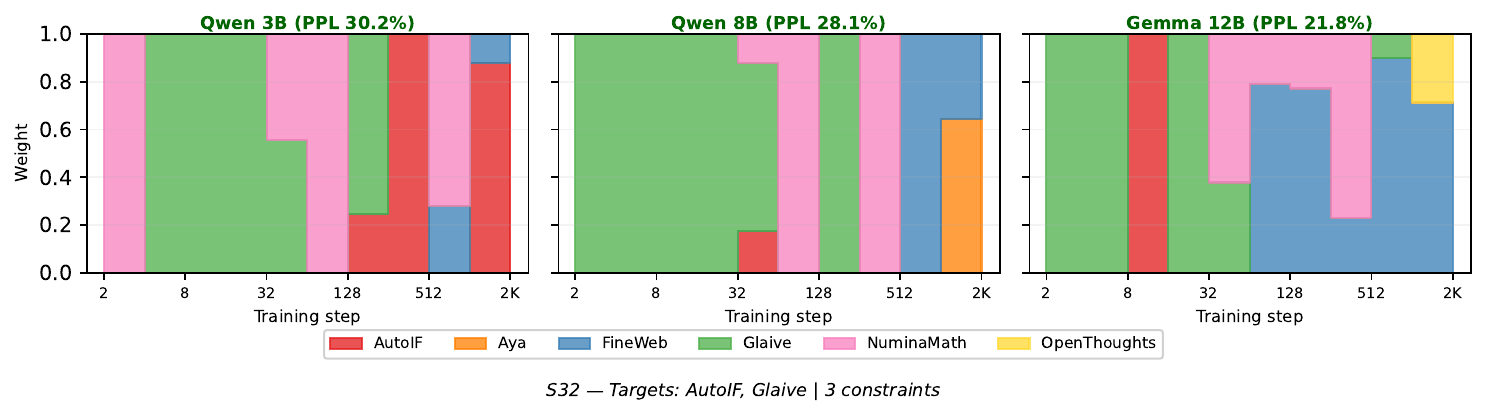}
    \includegraphics[width=\textwidth]{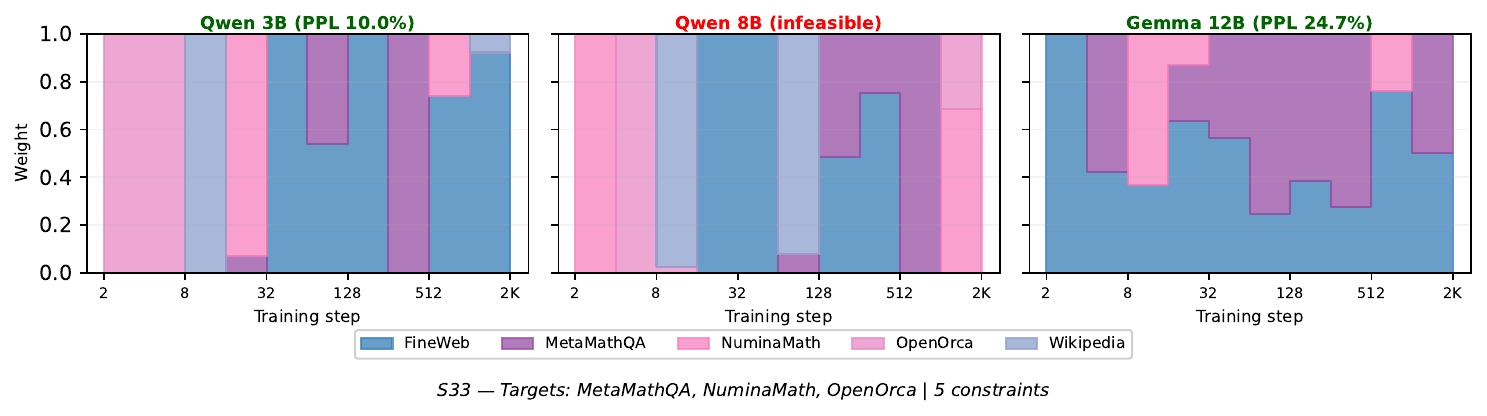}
    \includegraphics[width=\textwidth]{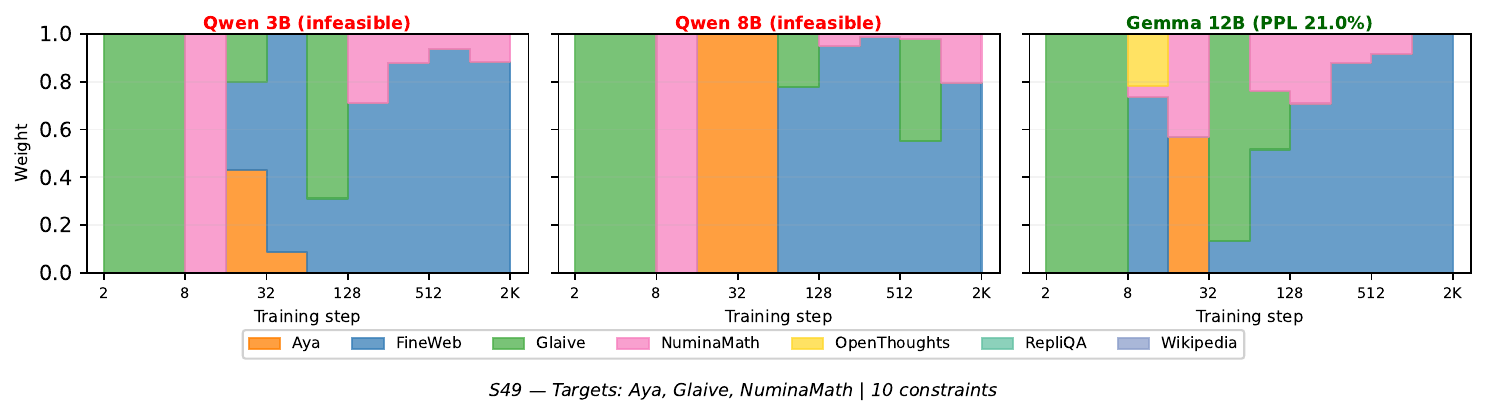}
     \vspace{-0.2cm}
    \caption{Evolution of dataset weights during training under a schedule of updates with dense warmup, for Qwen 3B, Qwen 8B, and Gemma 12B at LoRA rank~32. Weights are piecewise constant between controller recomputation steps.}

    \label{fig:weight_evolution}
\end{figure}

Figure~\ref{fig:weight_evolution} illustrates how the dynamic weight allocation varies across model architectures for different scenarios.  In S28 (7 constraints), all three models first emphasize the target domains, with Glaive dominating steps~2--16, before shifting weight towards FineWeb, which acts as a regularizer, later in training. In S32 (3 constraints) models converge to similar allocation strategies early on, but differ in a later phase. S33 resembles S28: weight allocations diverge across models, though all place larger weights on FineWeb in later stages. Finally, S49 is a highly constrained scenario (10 constraints) where only Gemma3-12B finds a feasible solution. While the three models begin with the same strategy, Gemma3-12B diverges at step~8, gradually settling into a stable FineWeb-dominated allocation, effectively leveraging a general-purpose data source to satisfy all constraints simultaneously while still reducing perplexity on the target domains by 21\%.